%% file: main.tex
\documentclass[runningheads]{llncs}
\usepackage{graphicx}
\usepackage{amsmath,amssymb} 
\graphicspath{{./figures/}}
\usepackage{color}
\usepackage{multirow}
\usepackage{array}
\usepackage[final]{pdfpages}
\usepackage[width=122mm,left=12mm,paperwidth=146mm,height=193mm,top=12mm,paperheight=217mm]{geometry}

\newcommand\alignvspace{-0.5cm}
\newcommand\figurevspace{-0.5cm}
\newcommand\tablevspace{-0.5cm}
\newcolumntype{M}{>{$\vcenter\bgroup\hbox\bgroup}c<{\egroup\egroup$}}

\begin{document}
\pagestyle{headings}
\mainmatter
\def\ECCV16SubNumber{1762}  

\title{3D-R2N2: A Unified Approach for Single and Multi-view 3D Object Reconstruction} 
\titlerunning{3D-R2N2}

\iftrue
\author{Christopher B. Choy \enskip Danfei Xu\thanks{indicates equal contribution.} \enskip JunYoung Gwak$^\star$ \enskip \\ Kevin Chen \enskip Silvio Savarese}
\authorrunning{C. B. Choy, D. Xu, J. Gwak, K. Chen, and S. Savarese}
\institute{Stanford University \\ \email{\{chrischoy, danfei, jgwak, kchen92, ssilvio\}@stanford.edu}}
\fi

\maketitle

\input{abstract.tex}
\input{intro}

\input{rnn}
\input{network}
\input{implementation}
\input{experiments}
\input{conclusion}
\input{acknowledgement}

\bibliographystyle{splncs}
\bibliography{references}

\end{document}


\pagestyle{headings}
\mainmatter
\def\ECCV16SubNumber{1762}  

\title{Supplementary Material for the Paper ``3D-R2N2: A Unified Approach for Single and Multi-view 3D Object Reconstruction``} 

\titlerunning{3D-R2N2}
\authorrunning{C. B. Choy, D. Xu, J. Gwak, K. Chen, and S. Savarese}

\author{}
\institute{}

\maketitle

\section{Single-View Real-World Image Reconstruction}

In this section, we present more single-view reconstruction results using the PASCAL VOC 2012 dataset and its corresponding 3D models from the PASCAL 3D+ dataset. Please refer to the main paper for more details.

\begin{figure}[htp!]
    \centering
    \begin{tabular}{ccccccccc}
        &\tiny{Input} & \tiny{Ground Truth} & \tiny{Ours} & \tiny{Kar et al.} & \tiny{Input} & \tiny{Ground Truth} & \tiny{Ours} & \tiny{Kar et al.}\\
      &\includegraphics[height=\wx\linewidth,width=\wx\linewidth,keepaspectratio]{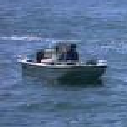} &  
      \includegraphics[height=\wx\linewidth,width=\wx\linewidth,keepaspectratio]{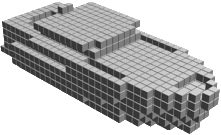} & 
      \includegraphics[height=\wx\linewidth,width=\wx\linewidth,keepaspectratio]{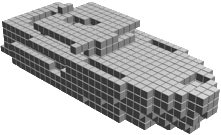} &   
      \includegraphics[height=\wx\linewidth,width=\wx\linewidth,keepaspectratio]{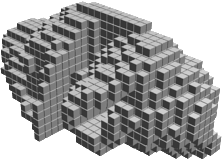} & 
      \includegraphics[height=\wx\linewidth,width=\wx\linewidth,keepaspectratio]{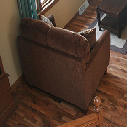} &
      \includegraphics[height=\wx\linewidth,width=\wx\linewidth,keepaspectratio]{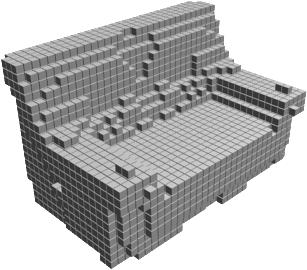} &   
      \includegraphics[height=\wx\linewidth,width=\wx\linewidth,keepaspectratio]{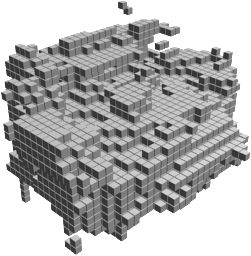} & 
      \includegraphics[height=\wx\linewidth,width=\wx\linewidth,keepaspectratio]{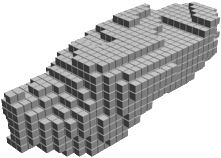} \\
      
      &\includegraphics[height=\wx\linewidth,width=\wx\linewidth,keepaspectratio]{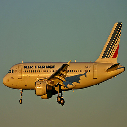} &  
      \includegraphics[height=\wx\linewidth,width=\wx\linewidth,keepaspectratio]{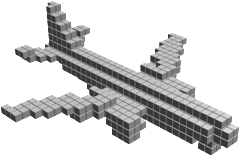} & 
      \includegraphics[height=\wx\linewidth,width=\wx\linewidth,keepaspectratio]{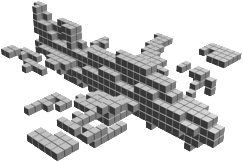} &   
      \includegraphics[height=\wx\linewidth,width=\wx\linewidth,keepaspectratio]{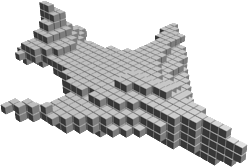} & 
      \includegraphics[height=\wx\linewidth,width=\wx\linewidth,keepaspectratio]{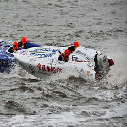} &
      \includegraphics[height=\wx\linewidth,width=\wx\linewidth,keepaspectratio]{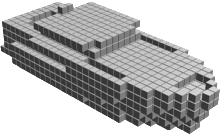} &   
      \includegraphics[height=\wx\linewidth,width=\wx\linewidth,keepaspectratio]{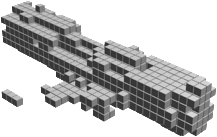} & 
      \includegraphics[height=\wx\linewidth,width=\wx\linewidth,keepaspectratio]{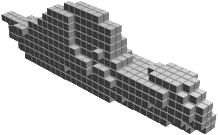} \\
      
      &\includegraphics[height=\wx\linewidth,width=\wx\linewidth,keepaspectratio]{figures/pascal3d_appendix/2008_000016_1_input.png} &  
      \includegraphics[height=\wx\linewidth,width=\wx\linewidth,keepaspectratio]{figures/pascal3d_appendix/2008_000016_1_gt.png} & 
      \includegraphics[height=\wx\linewidth,width=\wx\linewidth,keepaspectratio]{figures/pascal3d_appendix/2008_000016_1_ours.png} &   
      \includegraphics[height=\wx\linewidth,width=\wx\linewidth,keepaspectratio]{figures/pascal3d_appendix/2008_000016_1_cs.png} & 
      \includegraphics[height=\wx\linewidth,width=\wx\linewidth,keepaspectratio]{figures/pascal3d_appendix/2009_004108_1_input.png} &
      \includegraphics[height=\wx\linewidth,width=\wx\linewidth,keepaspectratio]{figures/pascal3d_appendix/2009_004108_1_gt.png} &   
      \includegraphics[height=\wx\linewidth,width=\wx\linewidth,keepaspectratio]{figures/pascal3d_appendix/2009_004108_1_ours.png} & 
      \includegraphics[height=\wx\linewidth,width=\wx\linewidth,keepaspectratio]{figures/pascal3d_appendix/2009_004108_1_cs.png} \\
      
      &\includegraphics[height=\wx\linewidth,width=\wx\linewidth,keepaspectratio]{figures/pascal3d_appendix/2010_000295_1_input.png} &  
      \includegraphics[height=\wx\linewidth,width=\wx\linewidth,keepaspectratio]{figures/pascal3d_appendix/2010_000295_1_gt.png} & 
      \includegraphics[height=\wx\linewidth,width=\wx\linewidth,keepaspectratio]{figures/pascal3d_appendix/2010_000295_1_ours.png} &   
      \includegraphics[height=\wx\linewidth,width=\wx\linewidth,keepaspectratio]{figures/pascal3d_appendix/2010_000295_1_cs.png} & 
      \includegraphics[height=\wx\linewidth,width=\wx\linewidth,keepaspectratio]{figures/pascal3d_appendix/2010_000573_1_input.png} &
      \includegraphics[height=\wx\linewidth,width=\wx\linewidth,keepaspectratio]{figures/pascal3d_appendix/2010_000573_1_gt.png} &   
      \includegraphics[height=\wx\linewidth,width=\wx\linewidth,keepaspectratio]{figures/pascal3d_appendix/2010_000573_1_ours.png} & 
      \includegraphics[height=\wx\linewidth,width=\wx\linewidth,keepaspectratio]{figures/pascal3d_appendix/2010_000573_1_cs.png} \\
      
      &\includegraphics[height=\wx\linewidth,width=\wx\linewidth,keepaspectratio]{figures/pascal3d_appendix/2010_000927_1_input.png} &  
      \includegraphics[height=\wx\linewidth,width=\wx\linewidth,keepaspectratio]{figures/pascal3d_appendix/2010_000927_1_gt.png} & 
      \includegraphics[height=\wx\linewidth,width=\wx\linewidth,keepaspectratio]{figures/pascal3d_appendix/2010_000927_1_ours.png} &   
      \includegraphics[height=\wx\linewidth,width=\wx\linewidth,keepaspectratio]{figures/pascal3d_appendix/2010_000927_1_cs.png} & 
      \includegraphics[height=\wx\linewidth,width=\wx\linewidth,keepaspectratio]{figures/pascal3d_appendix/2010_001525_1_input.png} &
      \includegraphics[height=\wx\linewidth,width=\wx\linewidth,keepaspectratio]{figures/pascal3d_appendix/2010_001525_1_gt.png} &   
      \includegraphics[height=\wx\linewidth,width=\wx\linewidth,keepaspectratio]{figures/pascal3d_appendix/2010_001525_1_ours.png} & 
      \includegraphics[height=\wx\linewidth,width=\wx\linewidth,keepaspectratio]{figures/pascal3d_appendix/2010_001525_1_cs.png} \\
      
      &\includegraphics[height=\wx\linewidth,width=\wx\linewidth,keepaspectratio]{figures/pascal3d_appendix/2010_001967_2_input.png} &  
      \includegraphics[height=\wx\linewidth,width=\wx\linewidth,keepaspectratio]{figures/pascal3d_appendix/2010_001967_2_gt.png} & 
      \includegraphics[height=\wx\linewidth,width=\wx\linewidth,keepaspectratio]{figures/pascal3d_appendix/2010_001967_2_ours.png} &   
      \includegraphics[height=\wx\linewidth,width=\wx\linewidth,keepaspectratio]{figures/pascal3d_appendix/2010_001967_2_cs.png} & 
      \includegraphics[height=\wx\linewidth,width=\wx\linewidth,keepaspectratio]{figures/pascal3d_appendix/2008_001040_1_input.png} &
      \includegraphics[height=\wx\linewidth,width=\wx\linewidth,keepaspectratio]{figures/pascal3d_appendix/2008_001040_1_gt.png} &   
      \includegraphics[height=\wx\linewidth,width=\wx\linewidth,keepaspectratio]{figures/pascal3d_appendix/2008_001040_1_ours.png} & 
      \includegraphics[height=\wx\linewidth,width=\wx\linewidth,keepaspectratio]{figures/pascal3d_appendix/2008_001040_1_cs.png} \\
      
      &\includegraphics[height=\wx\linewidth,width=\wx\linewidth,keepaspectratio]{figures/pascal3d_appendix/2009_005239_3_input.png} &  
      \includegraphics[height=\wx\linewidth,width=\wx\linewidth,keepaspectratio]{figures/pascal3d_appendix/2009_005239_3_gt.png} & 
      \includegraphics[height=\wx\linewidth,width=\wx\linewidth,keepaspectratio]{figures/pascal3d_appendix/2009_005239_3_ours.png} &   
      \includegraphics[height=\wx\linewidth,width=\wx\linewidth,keepaspectratio]{figures/pascal3d_appendix/2009_005239_3_cs.png} & 
      \includegraphics[height=\wx\linewidth,width=\wx\linewidth,keepaspectratio]{figures/pascal3d_appendix/2010_000342_1_input.png} &
      \includegraphics[height=\wx\linewidth,width=\wx\linewidth,keepaspectratio]{figures/pascal3d_appendix/2010_000342_1_gt.png} &   
      \includegraphics[height=\wx\linewidth,width=\wx\linewidth,keepaspectratio]{figures/pascal3d_appendix/2010_000342_1_ours.png} & 
      \includegraphics[height=\wx\linewidth,width=\wx\linewidth,keepaspectratio]{figures/pascal3d_appendix/2010_000342_1_cs.png} \\
      
      &\includegraphics[height=\wx\linewidth,width=\wx\linewidth,keepaspectratio]{figures/pascal3d_appendix/2010_000622_1_input.png} &  
      \includegraphics[height=\wx\linewidth,width=\wx\linewidth,keepaspectratio]{figures/pascal3d_appendix/2010_000622_1_gt.png} & 
      \includegraphics[height=\wx\linewidth,width=\wx\linewidth,keepaspectratio]{figures/pascal3d_appendix/2010_000622_1_ours.png} &   
      \includegraphics[height=\wx\linewidth,width=\wx\linewidth,keepaspectratio]{figures/pascal3d_appendix/2010_000622_1_cs.png} & 
      \includegraphics[height=\wx\linewidth,width=\wx\linewidth,keepaspectratio]{figures/pascal3d_appendix/2010_000952_1_input.png} &
      \includegraphics[height=\wx\linewidth,width=\wx\linewidth,keepaspectratio]{figures/pascal3d_appendix/2010_000952_1_gt.png} &   
      \includegraphics[height=\wx\linewidth,width=\wx\linewidth,keepaspectratio]{figures/pascal3d_appendix/2010_000952_1_ours.png} & 
      \includegraphics[height=\wx\linewidth,width=\wx\linewidth,keepaspectratio]{figures/pascal3d_appendix/2010_000952_1_cs.png} \\

    \end{tabular}
\end{figure}
      
\begin{figure}[htp!]
    \centering
    \begin{tabular}{ccccccccc}
        &\tiny{Input} & \tiny{Ground Truth} & \tiny{Ours} & \tiny{Kar et al.} & \tiny{Input} & \tiny{Ground Truth} & \tiny{Ours} & \tiny{Kar et al.}\\
      &\includegraphics[height=\wx\linewidth,width=\wx\linewidth,keepaspectratio]{figures/pascal3d_appendix/2010_001525_3_input.png} &  
      \includegraphics[height=\wx\linewidth,width=\wx\linewidth,keepaspectratio]{figures/pascal3d_appendix/2010_001525_3_gt.png} & 
      \includegraphics[height=\wx\linewidth,width=\wx\linewidth,keepaspectratio]{figures/pascal3d_appendix/2010_001525_3_ours.png} &   
      \includegraphics[height=\wx\linewidth,width=\wx\linewidth,keepaspectratio]{figures/pascal3d_appendix/2010_001525_3_cs.png} & 
      \includegraphics[height=\wx\linewidth,width=\wx\linewidth,keepaspectratio]{figures/pascal3d_appendix/2010_003981_2_input.png} &
      \includegraphics[height=\wx\linewidth,width=\wx\linewidth,keepaspectratio]{figures/pascal3d_appendix/2010_003981_2_gt.png} &   
      \includegraphics[height=\wx\linewidth,width=\wx\linewidth,keepaspectratio]{figures/pascal3d_appendix/2010_003981_2_ours.png} & 
      \includegraphics[height=\wx\linewidth,width=\wx\linewidth,keepaspectratio]{figures/pascal3d_appendix/2010_003981_2_cs.png} \\

      &\includegraphics[height=\wx\linewidth,width=\wx\linewidth,keepaspectratio]{figures/pascal3d_appendix/2008_002926_1_input.png} &  
      \includegraphics[height=\wx\linewidth,width=\wx\linewidth,keepaspectratio]{figures/pascal3d_appendix/2008_002926_1_gt.png} & 
      \includegraphics[height=\wx\linewidth,width=\wx\linewidth,keepaspectratio]{figures/pascal3d_appendix/2008_002926_1_ours.png} &   
      \includegraphics[height=\wx\linewidth,width=\wx\linewidth,keepaspectratio]{figures/pascal3d_appendix/2008_002926_1_cs.png} & 
      \includegraphics[height=\wx\linewidth,width=\wx\linewidth,keepaspectratio]{figures/pascal3d_appendix/2009_005257_2_input.png} &
      \includegraphics[height=\wx\linewidth,width=\wx\linewidth,keepaspectratio]{figures/pascal3d_appendix/2009_005257_2_gt.png} &   
      \includegraphics[height=\wx\linewidth,width=\wx\linewidth,keepaspectratio]{figures/pascal3d_appendix/2009_005257_2_ours.png} & 
      \includegraphics[height=\wx\linewidth,width=\wx\linewidth,keepaspectratio]{figures/pascal3d_appendix/2009_005257_2_cs.png} \\
      
      &\includegraphics[height=\wx\linewidth,width=\wx\linewidth,keepaspectratio]{figures/pascal3d_appendix/2010_000381_1_input.png} &  
      \includegraphics[height=\wx\linewidth,width=\wx\linewidth,keepaspectratio]{figures/pascal3d_appendix/2010_000381_1_gt.png} & 
      \includegraphics[height=\wx\linewidth,width=\wx\linewidth,keepaspectratio]{figures/pascal3d_appendix/2010_000381_1_ours.png} &   
      \includegraphics[height=\wx\linewidth,width=\wx\linewidth,keepaspectratio]{figures/pascal3d_appendix/2010_000381_1_cs.png} & 
      \includegraphics[height=\wx\linewidth,width=\wx\linewidth,keepaspectratio]{figures/pascal3d_appendix/2010_000705_1_input.png} &
      \includegraphics[height=\wx\linewidth,width=\wx\linewidth,keepaspectratio]{figures/pascal3d_appendix/2010_000705_1_gt.png} &   
      \includegraphics[height=\wx\linewidth,width=\wx\linewidth,keepaspectratio]{figures/pascal3d_appendix/2010_000705_1_ours.png} & 
      \includegraphics[height=\wx\linewidth,width=\wx\linewidth,keepaspectratio]{figures/pascal3d_appendix/2010_000705_1_cs.png} \\
      
      &\includegraphics[height=\wx\linewidth,width=\wx\linewidth,keepaspectratio]{figures/pascal3d_appendix/2010_001049_2_input.png} &  
      \includegraphics[height=\wx\linewidth,width=\wx\linewidth,keepaspectratio]{figures/pascal3d_appendix/2010_001049_2_gt.png} & 
      \includegraphics[height=\wx\linewidth,width=\wx\linewidth,keepaspectratio]{figures/pascal3d_appendix/2010_001049_2_ours.png} &   
      \includegraphics[height=\wx\linewidth,width=\wx\linewidth,keepaspectratio]{figures/pascal3d_appendix/2010_001049_2_cs.png} & 
      \includegraphics[height=\wx\linewidth,width=\wx\linewidth,keepaspectratio]{figures/pascal3d_appendix/2010_001574_2_input.png} &
      \includegraphics[height=\wx\linewidth,width=\wx\linewidth,keepaspectratio]{figures/pascal3d_appendix/2010_001574_2_gt.png} &   
      \includegraphics[height=\wx\linewidth,width=\wx\linewidth,keepaspectratio]{figures/pascal3d_appendix/2010_001574_2_ours.png} & 
      \includegraphics[height=\wx\linewidth,width=\wx\linewidth,keepaspectratio]{figures/pascal3d_appendix/2010_001574_2_cs.png} \\
      
      &\includegraphics[height=\wx\linewidth,width=\wx\linewidth,keepaspectratio]{figures/pascal3d_appendix/2010_005192_1_input.png} &  
      \includegraphics[height=\wx\linewidth,width=\wx\linewidth,keepaspectratio]{figures/pascal3d_appendix/2010_005192_1_gt.png} & 
      \includegraphics[height=\wx\linewidth,width=\wx\linewidth,keepaspectratio]{figures/pascal3d_appendix/2010_005192_1_ours.png} &   
      \includegraphics[height=\wx\linewidth,width=\wx\linewidth,keepaspectratio]{figures/pascal3d_appendix/2010_005192_1_cs.png} & 
      \includegraphics[height=\wx\linewidth,width=\wx\linewidth,keepaspectratio]{figures/pascal3d_appendix/2009_003571_1_input.png} &
      \includegraphics[height=\wx\linewidth,width=\wx\linewidth,keepaspectratio]{figures/pascal3d_appendix/2009_003571_1_gt.png} &   
      \includegraphics[height=\wx\linewidth,width=\wx\linewidth,keepaspectratio]{figures/pascal3d_appendix/2009_003571_1_ours.png} & 
      \includegraphics[height=\wx\linewidth,width=\wx\linewidth,keepaspectratio]{figures/pascal3d_appendix/2009_003571_1_cs.png} \\
      
      &\includegraphics[height=\wx\linewidth,width=\wx\linewidth,keepaspectratio]{figures/pascal3d_appendix/2009_005279_1_input.png} &  
      \includegraphics[height=\wx\linewidth,width=\wx\linewidth,keepaspectratio]{figures/pascal3d_appendix/2009_005279_1_gt.png} & 
      \includegraphics[height=\wx\linewidth,width=\wx\linewidth,keepaspectratio]{figures/pascal3d_appendix/2009_005279_1_ours.png} &   
      \includegraphics[height=\wx\linewidth,width=\wx\linewidth,keepaspectratio]{figures/pascal3d_appendix/2009_005279_1_cs.png} & 
      \includegraphics[height=\wx\linewidth,width=\wx\linewidth,keepaspectratio]{figures/pascal3d_appendix/2010_000449_1_input.png} &
      \includegraphics[height=\wx\linewidth,width=\wx\linewidth,keepaspectratio]{figures/pascal3d_appendix/2010_000449_1_gt.png} &   
      \includegraphics[height=\wx\linewidth,width=\wx\linewidth,keepaspectratio]{figures/pascal3d_appendix/2010_000449_1_ours.png} & 
      \includegraphics[height=\wx\linewidth,width=\wx\linewidth,keepaspectratio]{figures/pascal3d_appendix/2010_000449_1_cs.png} \\
      
      &\includegraphics[height=\wx\linewidth,width=\wx\linewidth,keepaspectratio]{figures/pascal3d_appendix/2010_000898_1_input.png} &  
      \includegraphics[height=\wx\linewidth,width=\wx\linewidth,keepaspectratio]{figures/pascal3d_appendix/2010_000898_1_gt.png} & 
      \includegraphics[height=\wx\linewidth,width=\wx\linewidth,keepaspectratio]{figures/pascal3d_appendix/2010_000898_1_ours.png} &   
      \includegraphics[height=\wx\linewidth,width=\wx\linewidth,keepaspectratio]{figures/pascal3d_appendix/2010_000898_1_cs.png} & 
      \includegraphics[height=\wx\linewidth,width=\wx\linewidth,keepaspectratio]{figures/pascal3d_appendix/2010_001080_1_input.png} &
      \includegraphics[height=\wx\linewidth,width=\wx\linewidth,keepaspectratio]{figures/pascal3d_appendix/2010_001080_1_gt.png} &   
      \includegraphics[height=\wx\linewidth,width=\wx\linewidth,keepaspectratio]{figures/pascal3d_appendix/2010_001080_1_ours.png} & 
      \includegraphics[height=\wx\linewidth,width=\wx\linewidth,keepaspectratio]{figures/pascal3d_appendix/2010_001080_1_cs.png} \\
      
      &\includegraphics[height=\wx\linewidth,width=\wx\linewidth,keepaspectratio]{figures/pascal3d_appendix/2010_001752_1_input.png} &  
      \includegraphics[height=\wx\linewidth,width=\wx\linewidth,keepaspectratio]{figures/pascal3d_appendix/2010_001752_1_gt.png} & 
      \includegraphics[height=\wx\linewidth,width=\wx\linewidth,keepaspectratio]{figures/pascal3d_appendix/2010_001752_1_ours.png} &   
      \includegraphics[height=\wx\linewidth,width=\wx\linewidth,keepaspectratio]{figures/pascal3d_appendix/2010_001752_1_cs.png} & 
      \includegraphics[height=\wx\linewidth,width=\wx\linewidth,keepaspectratio]{figures/pascal3d_appendix/2010_005493_1_input.png} &
      \includegraphics[height=\wx\linewidth,width=\wx\linewidth,keepaspectratio]{figures/pascal3d_appendix/2010_005493_1_gt.png} &   
      \includegraphics[height=\wx\linewidth,width=\wx\linewidth,keepaspectratio]{figures/pascal3d_appendix/2010_005493_1_ours.png} & 
      \includegraphics[height=\wx\linewidth,width=\wx\linewidth,keepaspectratio]{figures/pascal3d_appendix/2010_005493_1_cs.png} \\
      
      &\includegraphics[height=\wx\linewidth,width=\wx\linewidth,keepaspectratio]{figures/pascal3d_appendix/2009_003884_1_input.png} &  
      \includegraphics[height=\wx\linewidth,width=\wx\linewidth,keepaspectratio]{figures/pascal3d_appendix/2009_003884_1_gt.png} & 
      \includegraphics[height=\wx\linewidth,width=\wx\linewidth,keepaspectratio]{figures/pascal3d_appendix/2009_003884_1_ours.png} &   
      \includegraphics[height=\wx\linewidth,width=\wx\linewidth,keepaspectratio]{figures/pascal3d_appendix/2009_003884_1_cs.png} & 
      \includegraphics[height=\wx\linewidth,width=\wx\linewidth,keepaspectratio]{figures/pascal3d_appendix/2010_000272_1_input.png} &
      \includegraphics[height=\wx\linewidth,width=\wx\linewidth,keepaspectratio]{figures/pascal3d_appendix/2010_000272_1_gt.png} &   
      \includegraphics[height=\wx\linewidth,width=\wx\linewidth,keepaspectratio]{figures/pascal3d_appendix/2010_000272_1_ours.png} & 
      \includegraphics[height=\wx\linewidth,width=\wx\linewidth,keepaspectratio]{figures/pascal3d_appendix/2010_000272_1_cs.png} \\
      
      &\includegraphics[height=\wx\linewidth,width=\wx\linewidth,keepaspectratio]{figures/pascal3d_appendix/2010_000510_1_input.png} &  
      \includegraphics[height=\wx\linewidth,width=\wx\linewidth,keepaspectratio]{figures/pascal3d_appendix/2010_000510_1_gt.png} & 
      \includegraphics[height=\wx\linewidth,width=\wx\linewidth,keepaspectratio]{figures/pascal3d_appendix/2010_000510_1_ours.png} &   
      \includegraphics[height=\wx\linewidth,width=\wx\linewidth,keepaspectratio]{figures/pascal3d_appendix/2010_000510_1_cs.png} & 
      \includegraphics[height=\wx\linewidth,width=\wx\linewidth,keepaspectratio]{figures/pascal3d_appendix/2010_000906_1_input.png} &
      \includegraphics[height=\wx\linewidth,width=\wx\linewidth,keepaspectratio]{figures/pascal3d_appendix/2010_000906_1_gt.png} &   
      \includegraphics[height=\wx\linewidth,width=\wx\linewidth,keepaspectratio]{figures/pascal3d_appendix/2010_000906_1_ours.png} & 
      \includegraphics[height=\wx\linewidth,width=\wx\linewidth,keepaspectratio]{figures/pascal3d_appendix/2010_000906_1_cs.png} \\
      
      &\includegraphics[height=\wx\linewidth,width=\wx\linewidth,keepaspectratio]{figures/pascal3d_appendix/2010_001246_2_input.png} &  
      \includegraphics[height=\wx\linewidth,width=\wx\linewidth,keepaspectratio]{figures/pascal3d_appendix/2010_001246_2_gt.png} & 
      \includegraphics[height=\wx\linewidth,width=\wx\linewidth,keepaspectratio]{figures/pascal3d_appendix/2010_001246_2_ours.png} &   
      \includegraphics[height=\wx\linewidth,width=\wx\linewidth,keepaspectratio]{figures/pascal3d_appendix/2010_001246_2_cs.png} & 
      \includegraphics[height=\wx\linewidth,width=\wx\linewidth,keepaspectratio]{figures/pascal3d_appendix/2010_001763_1_input.png} &
      \includegraphics[height=\wx\linewidth,width=\wx\linewidth,keepaspectratio]{figures/pascal3d_appendix/2010_001763_1_gt.png} &   
      \includegraphics[height=\wx\linewidth,width=\wx\linewidth,keepaspectratio]{figures/pascal3d_appendix/2010_001763_1_ours.png} & 
      \includegraphics[height=\wx\linewidth,width=\wx\linewidth,keepaspectratio]{figures/pascal3d_appendix/2010_001763_1_cs.png} \\

    \end{tabular}
    \caption{Further reconstruction samples of PASCAL VOC 2012 dataset.}
\end{figure}

\newpage
\section{Online Product Dataset Reconstruction}

In this section, we present more multi-view reconstruction results using the Online Product dataset. Please refer to the main paper for more details.

\begin{figure}[htp!]
\begin{tabular}{cccccccccc}
  \includegraphics[width=\wxp\linewidth,height=\wxp\linewidth,keepaspectratio]{figures/appendix/361123986585/0} &  
  \includegraphics[width=\wxp\linewidth,height=\wxp\linewidth,keepaspectratio]{figures/appendix/361123986585/1} & 
  \includegraphics[width=\wxp\linewidth,height=\wxp\linewidth,keepaspectratio]{figures/appendix/141495820360/0} &   
  \includegraphics[width=\wxp\linewidth,height=\wxp\linewidth,keepaspectratio]{figures/appendix/141495820360/1} &
  \includegraphics[width=\wxp\linewidth,height=\wxp\linewidth,keepaspectratio]{figures/appendix/331390339252/0} &
  \includegraphics[width=\wxp\linewidth,height=\wxp\linewidth,keepaspectratio]{figures/appendix/331390339252/1} &
  \includegraphics[width=\wxp\linewidth,height=\wxp\linewidth,keepaspectratio]{figures/appendix/331390339252/2} &
  \includegraphics[width=\wxp\linewidth,height=\wxp\linewidth,keepaspectratio]{figures/appendix/400683697826/0} &
  \includegraphics[width=\wxp\linewidth,height=\wxp\linewidth,keepaspectratio]{figures/appendix/400683697826/1} &
  \includegraphics[width=\wxp\linewidth,height=\wxp\linewidth,keepaspectratio]{figures/appendix/400683697826/2} \\
  \includegraphics[width=\wxp\linewidth,height=\wxp\linewidth,keepaspectratio]{figures/appendix/361123986585/0_pred} &  
  \includegraphics[width=\wxp\linewidth,height=\wxp\linewidth,keepaspectratio]{figures/appendix/361123986585/1_pred} & 
  \includegraphics[width=\wxp\linewidth,height=\wxp\linewidth,keepaspectratio]{figures/appendix/141495820360/0_pred} &   
  \includegraphics[width=\wxp\linewidth,height=\wxp\linewidth,keepaspectratio]{figures/appendix/141495820360/1_pred} &
  \includegraphics[width=\wxp\linewidth,height=\wxp\linewidth,keepaspectratio]{figures/appendix/331390339252/0_pred} &   
  \includegraphics[width=\wxp\linewidth,height=\wxp\linewidth,keepaspectratio]{figures/appendix/331390339252/1_pred} &
  \includegraphics[width=\wxp\linewidth,height=\wxp\linewidth,keepaspectratio]{figures/appendix/331390339252/2_pred} &   
  \includegraphics[width=\wxp\linewidth,height=\wxp\linewidth,keepaspectratio]{figures/appendix/400683697826/0_pred} &   
  \includegraphics[width=\wxp\linewidth,height=\wxp\linewidth,keepaspectratio]{figures/appendix/400683697826/1_pred} &
  \includegraphics[width=\wxp\linewidth,height=\wxp\linewidth,keepaspectratio]{figures/appendix/400683697826/2_pred} \\

  \includegraphics[width=\wxp\linewidth,height=\wxp\linewidth,keepaspectratio]{figures/appendix/141081571425/0} &  
  \includegraphics[width=\wxp\linewidth,height=\wxp\linewidth,keepaspectratio]{figures/appendix/141081571425/1} & 
  \includegraphics[width=\wxp\linewidth,height=\wxp\linewidth,keepaspectratio]{figures/appendix/141081571425/2} &   
  \includegraphics[width=\wxp\linewidth,height=\wxp\linewidth,keepaspectratio]{figures/appendix/141081571425/3} &
  \includegraphics[width=\wxp\linewidth,height=\wxp\linewidth,keepaspectratio]{figures/appendix/141456351141/0} &
  \includegraphics[width=\wxp\linewidth,height=\wxp\linewidth,keepaspectratio]{figures/appendix/141456351141/1} &
  \includegraphics[width=\wxp\linewidth,height=\wxp\linewidth,keepaspectratio]{figures/appendix/141456351141/2} &
  \includegraphics[width=\wxp\linewidth,height=\wxp\linewidth,keepaspectratio]{figures/appendix/331293683348/0} &
  \includegraphics[width=\wxp\linewidth,height=\wxp\linewidth,keepaspectratio]{figures/appendix/331293683348/1} &
  \includegraphics[width=\wxp\linewidth,height=\wxp\linewidth,keepaspectratio]{figures/appendix/331293683348/2} \\
  \includegraphics[width=\wxp\linewidth,height=\wxp\linewidth,keepaspectratio]{figures/appendix/141081571425/0_pred} &  
  \includegraphics[width=\wxp\linewidth,height=\wxp\linewidth,keepaspectratio]{figures/appendix/141081571425/1_pred} & 
  \includegraphics[width=\wxp\linewidth,height=\wxp\linewidth,keepaspectratio]{figures/appendix/141081571425/2_pred} &   
  \includegraphics[width=\wxp\linewidth,height=\wxp\linewidth,keepaspectratio]{figures/appendix/141081571425/3_pred} &
  \includegraphics[width=\wxp\linewidth,height=\wxp\linewidth,keepaspectratio]{figures/appendix/141456351141/0_pred} &   
  \includegraphics[width=\wxp\linewidth,height=\wxp\linewidth,keepaspectratio]{figures/appendix/141456351141/1_pred} &
  \includegraphics[width=\wxp\linewidth,height=\wxp\linewidth,keepaspectratio]{figures/appendix/141456351141/2_pred} &   
  \includegraphics[width=\wxp\linewidth,height=\wxp\linewidth,keepaspectratio]{figures/appendix/331293683348/0_pred} &   
  \includegraphics[width=\wxp\linewidth,height=\wxp\linewidth,keepaspectratio]{figures/appendix/331293683348/1_pred} &
  \includegraphics[width=\wxp\linewidth,height=\wxp\linewidth,keepaspectratio]{figures/appendix/331293683348/2_pred} \\
  
  \includegraphics[width=\wxp\linewidth,height=\wxp\linewidth,keepaspectratio]{figures/appendix/_111492871577/0} &  
  \includegraphics[width=\wxp\linewidth,height=\wxp\linewidth,keepaspectratio]{figures/appendix/_111492871577/1} & 
  \includegraphics[width=\wxp\linewidth,height=\wxp\linewidth,keepaspectratio]{figures/appendix/_121649600790/0} &   
  \includegraphics[width=\wxp\linewidth,height=\wxp\linewidth,keepaspectratio]{figures/appendix/_121649600790/1} &
  \includegraphics[width=\wxp\linewidth,height=\wxp\linewidth,keepaspectratio]{figures/appendix/_111523466059/0} &
  \includegraphics[width=\wxp\linewidth,height=\wxp\linewidth,keepaspectratio]{figures/appendix/_111523466059/1} &
  \includegraphics[width=\wxp\linewidth,height=\wxp\linewidth,keepaspectratio]{figures/appendix/_111523466059/2} &
  \includegraphics[width=\wxp\linewidth,height=\wxp\linewidth,keepaspectratio]{figures/appendix/_111652483124/0} &
  \includegraphics[width=\wxp\linewidth,height=\wxp\linewidth,keepaspectratio]{figures/appendix/_111652483124/1} &
  \includegraphics[width=\wxp\linewidth,height=\wxp\linewidth,keepaspectratio]{figures/appendix/_111652483124/2} \\
  \includegraphics[width=\wxp\linewidth,height=\wxp\linewidth,keepaspectratio]{figures/appendix/_111492871577/0_pred} &  
  \includegraphics[width=\wxp\linewidth,height=\wxp\linewidth,keepaspectratio]{figures/appendix/_111492871577/1_pred} & 
  \includegraphics[width=\wxp\linewidth,height=\wxp\linewidth,keepaspectratio]{figures/appendix/_121649600790/0_pred} &   
  \includegraphics[width=\wxp\linewidth,height=\wxp\linewidth,keepaspectratio]{figures/appendix/_121649600790/1_pred} &
  \includegraphics[width=\wxp\linewidth,height=\wxp\linewidth,keepaspectratio]{figures/appendix/_111523466059/0_pred} &   
  \includegraphics[width=\wxp\linewidth,height=\wxp\linewidth,keepaspectratio]{figures/appendix/_111523466059/1_pred} &
  \includegraphics[width=\wxp\linewidth,height=\wxp\linewidth,keepaspectratio]{figures/appendix/_111523466059/2_pred} &   
  \includegraphics[width=\wxp\linewidth,height=\wxp\linewidth,keepaspectratio]{figures/appendix/_111652483124/0_pred} &   
  \includegraphics[width=\wxp\linewidth,height=\wxp\linewidth,keepaspectratio]{figures/appendix/_111652483124/1_pred} &
  \includegraphics[width=\wxp\linewidth,height=\wxp\linewidth,keepaspectratio]{figures/appendix/_111652483124/2_pred} \\
  
  \includegraphics[width=\wxp\linewidth,height=\wxp\linewidth,keepaspectratio]{figures/appendix/_121668598610/0} &  
  \includegraphics[width=\wxp\linewidth,height=\wxp\linewidth,keepaspectratio]{figures/appendix/_121668598610/1} & 
  \includegraphics[width=\wxp\linewidth,height=\wxp\linewidth,keepaspectratio]{figures/appendix/_121719768362/0} &   
  \includegraphics[width=\wxp\linewidth,height=\wxp\linewidth,keepaspectratio]{figures/appendix/_121719768362/1} &
  \includegraphics[width=\wxp\linewidth,height=\wxp\linewidth,keepaspectratio]{figures/appendix/_121019663343/0} &
  \includegraphics[width=\wxp\linewidth,height=\wxp\linewidth,keepaspectratio]{figures/appendix/_121019663343/1} &
  \includegraphics[width=\wxp\linewidth,height=\wxp\linewidth,keepaspectratio]{figures/appendix/_121019663343/2} &
  \includegraphics[width=\wxp\linewidth,height=\wxp\linewidth,keepaspectratio]{figures/appendix/_161481451447/0} &
  \includegraphics[width=\wxp\linewidth,height=\wxp\linewidth,keepaspectratio]{figures/appendix/_161481451447/1} &
  \includegraphics[width=\wxp\linewidth,height=\wxp\linewidth,keepaspectratio]{figures/appendix/_161481451447/2} \\
  \includegraphics[width=\wxp\linewidth,height=\wxp\linewidth,keepaspectratio]{figures/appendix/_121668598610/0_pred} &  
  \includegraphics[width=\wxp\linewidth,height=\wxp\linewidth,keepaspectratio]{figures/appendix/_121668598610/1_pred} & 
  \includegraphics[width=\wxp\linewidth,height=\wxp\linewidth,keepaspectratio]{figures/appendix/_121719768362/0_pred} &   
  \includegraphics[width=\wxp\linewidth,height=\wxp\linewidth,keepaspectratio]{figures/appendix/_121719768362/1_pred} &
  \includegraphics[width=\wxp\linewidth,height=\wxp\linewidth,keepaspectratio]{figures/appendix/_121019663343/0_pred} &   
  \includegraphics[width=\wxp\linewidth,height=\wxp\linewidth,keepaspectratio]{figures/appendix/_121019663343/1_pred} &
  \includegraphics[width=\wxp\linewidth,height=\wxp\linewidth,keepaspectratio]{figures/appendix/_121019663343/2_pred} &   
  \includegraphics[width=\wxp\linewidth,height=\wxp\linewidth,keepaspectratio]{figures/appendix/_161481451447/0_pred} &   
  \includegraphics[width=\wxp\linewidth,height=\wxp\linewidth,keepaspectratio]{figures/appendix/_161481451447/1_pred} &
  \includegraphics[width=\wxp\linewidth,height=\wxp\linewidth,keepaspectratio]{figures/appendix/_161481451447/2_pred} \\

  \includegraphics[width=\wxp\linewidth,height=\wxp\linewidth,keepaspectratio]{figures/appendix/_141494536080/0} &
  \includegraphics[width=\wxp\linewidth,height=\wxp\linewidth,keepaspectratio]{figures/appendix/_141494536080/1} &
  \includegraphics[width=\wxp\linewidth,height=\wxp\linewidth,keepaspectratio]{figures/appendix/_151502238166/0} &
  \includegraphics[width=\wxp\linewidth,height=\wxp\linewidth,keepaspectratio]{figures/appendix/_151502238166/1} &
  \includegraphics[width=\wxp\linewidth,height=\wxp\linewidth,keepaspectratio]{figures/appendix/_361123986588/0} &
  \includegraphics[width=\wxp\linewidth,height=\wxp\linewidth,keepaspectratio]{figures/appendix/_361123986588/1} &
    \includegraphics[width=\wxp\linewidth,height=\wxp\linewidth,keepaspectratio]{figures/appendix/_111530206799/0} &  
  \includegraphics[width=\wxp\linewidth,height=\wxp\linewidth,keepaspectratio]{figures/appendix/_111530206799/1} & 
  \includegraphics[width=\wxp\linewidth,height=\wxp\linewidth,keepaspectratio]{figures/appendix/_111530206799/2} &   
  \includegraphics[width=\wxp\linewidth,height=\wxp\linewidth,keepaspectratio]{figures/appendix/_111530206799/3} \\
  \includegraphics[width=\wxp\linewidth,height=\wxp\linewidth,keepaspectratio]{figures/appendix/_141494536080/0_pred} &   
  \includegraphics[width=\wxp\linewidth,height=\wxp\linewidth,keepaspectratio]{figures/appendix/_141494536080/1_pred} &
  \includegraphics[width=\wxp\linewidth,height=\wxp\linewidth,keepaspectratio]{figures/appendix/_151502238166/0_pred} &   
  \includegraphics[width=\wxp\linewidth,height=\wxp\linewidth,keepaspectratio]{figures/appendix/_151502238166/1_pred} &   
  \includegraphics[width=\wxp\linewidth,height=\wxp\linewidth,keepaspectratio]{figures/appendix/_361123986588/0_pred} &
  \includegraphics[width=\wxp\linewidth,height=\wxp\linewidth,keepaspectratio]{figures/appendix/_361123986588/1_pred} &
    \includegraphics[width=\wxp\linewidth,height=\wxp\linewidth,keepaspectratio]{figures/appendix/_111530206799/0_pred} &  
  \includegraphics[width=\wxp\linewidth,height=\wxp\linewidth,keepaspectratio]{figures/appendix/_111530206799/1_pred} & 
  \includegraphics[width=\wxp\linewidth,height=\wxp\linewidth,keepaspectratio]{figures/appendix/_111530206799/2_pred} &   
  \includegraphics[width=\wxp\linewidth,height=\wxp\linewidth,keepaspectratio]{figures/appendix/_111530206799/3_pred} \\
  
  \includegraphics[width=\wxp\linewidth,height=\wxp\linewidth,keepaspectratio]{figures/appendix/_351235098324/0} &  
  \includegraphics[width=\wxp\linewidth,height=\wxp\linewidth,keepaspectratio]{figures/appendix/_351235098324/1} & 
  \includegraphics[width=\wxp\linewidth,height=\wxp\linewidth,keepaspectratio]{figures/appendix/_351235098324/2} &   
  \includegraphics[width=\wxp\linewidth,height=\wxp\linewidth,keepaspectratio]{figures/appendix/_351235098324/3} &
  \includegraphics[width=\wxp\linewidth,height=\wxp\linewidth,keepaspectratio]{figures/appendix/_331389096582/0} &
  \includegraphics[width=\wxp\linewidth,height=\wxp\linewidth,keepaspectratio]{figures/appendix/_331389096582/1} &
  \includegraphics[width=\wxp\linewidth,height=\wxp\linewidth,keepaspectratio]{figures/appendix/_331389096582/2} &
  \includegraphics[width=\wxp\linewidth,height=\wxp\linewidth,keepaspectratio]{figures/appendix/_331389096582/3} &
  \includegraphics[width=\wxp\linewidth,height=\wxp\linewidth,keepaspectratio]{figures/appendix/_371195267608/0} &
  \includegraphics[width=\wxp\linewidth,height=\wxp\linewidth,keepaspectratio]{figures/appendix/_371195267608/1} \\
  \includegraphics[width=\wxp\linewidth,height=\wxp\linewidth,keepaspectratio]{figures/appendix/_351235098324/0_pred} &  
  \includegraphics[width=\wxp\linewidth,height=\wxp\linewidth,keepaspectratio]{figures/appendix/_351235098324/1_pred} & 
  \includegraphics[width=\wxp\linewidth,height=\wxp\linewidth,keepaspectratio]{figures/appendix/_351235098324/2_pred} &   
  \includegraphics[width=\wxp\linewidth,height=\wxp\linewidth,keepaspectratio]{figures/appendix/_351235098324/3_pred} &
  \includegraphics[width=\wxp\linewidth,height=\wxp\linewidth,keepaspectratio]{figures/appendix/_331389096582/0_pred} &   
  \includegraphics[width=\wxp\linewidth,height=\wxp\linewidth,keepaspectratio]{figures/appendix/_331389096582/1_pred} &
  \includegraphics[width=\wxp\linewidth,height=\wxp\linewidth,keepaspectratio]{figures/appendix/_331389096582/2_pred} &   
  \includegraphics[width=\wxp\linewidth,height=\wxp\linewidth,keepaspectratio]{figures/appendix/_331389096582/3_pred} &   
  \includegraphics[width=\wxp\linewidth,height=\wxp\linewidth,keepaspectratio]{figures/appendix/_371195267608/0_pred} &
  \includegraphics[width=\wxp\linewidth,height=\wxp\linewidth,keepaspectratio]{figures/appendix/_371195267608/1_pred} \\
\end{tabular}
\caption{Reconstructions of the Online Product dataset.}
\end{figure}

\begin{figure}[htp!]
\begin{tabular}{cccccccccc}
\includegraphics[width=\wxp\linewidth,height=\wxp\linewidth,keepaspectratio]{figures/appendix/10/0}  &
\includegraphics[width=\wxp\linewidth,height=\wxp\linewidth,keepaspectratio]{figures/appendix/10/1}  &
\includegraphics[width=\wxp\linewidth,height=\wxp\linewidth,keepaspectratio]{figures/appendix/10/2}  &
\includegraphics[width=\wxp\linewidth,height=\wxp\linewidth,keepaspectratio]{figures/appendix/10/3}  &
\includegraphics[width=\wxp\linewidth,height=\wxp\linewidth,keepaspectratio]{figures/appendix/10/4}  &
\includegraphics[width=\wxp\linewidth,height=\wxp\linewidth,keepaspectratio]{figures/appendix/12/0}  &
\includegraphics[width=\wxp\linewidth,height=\wxp\linewidth,keepaspectratio]{figures/appendix/12/1}  &
\includegraphics[width=\wxp\linewidth,height=\wxp\linewidth,keepaspectratio]{figures/appendix/12/2}  &
\includegraphics[width=\wxp\linewidth,height=\wxp\linewidth,keepaspectratio]{figures/appendix/12/3}  &
\includegraphics[width=\wxp\linewidth,height=\wxp\linewidth,keepaspectratio]{figures/appendix/12/4} \\ 

\includegraphics[width=\wxp\linewidth,height=\wxp\linewidth,keepaspectratio]{figures/appendix/10/0_pred}  &
\includegraphics[width=\wxp\linewidth,height=\wxp\linewidth,keepaspectratio]{figures/appendix/10/1_pred}  &
\includegraphics[width=\wxp\linewidth,height=\wxp\linewidth,keepaspectratio]{figures/appendix/10/2_pred}  &
\includegraphics[width=\wxp\linewidth,height=\wxp\linewidth,keepaspectratio]{figures/appendix/10/3_pred}  &
\includegraphics[width=\wxp\linewidth,height=\wxp\linewidth,keepaspectratio]{figures/appendix/10/4_pred}  &
\includegraphics[width=\wxp\linewidth,height=\wxp\linewidth,keepaspectratio]{figures/appendix/12/0_pred}  &
\includegraphics[width=\wxp\linewidth,height=\wxp\linewidth,keepaspectratio]{figures/appendix/12/1_pred}  &
\includegraphics[width=\wxp\linewidth,height=\wxp\linewidth,keepaspectratio]{figures/appendix/12/2_pred}  &
\includegraphics[width=\wxp\linewidth,height=\wxp\linewidth,keepaspectratio]{figures/appendix/12/3_pred}  &
\includegraphics[width=\wxp\linewidth,height=\wxp\linewidth,keepaspectratio]{figures/appendix/12/4_pred} \\ 

\includegraphics[width=\wxp\linewidth,height=\wxp\linewidth,keepaspectratio]{figures/appendix/57/0}  &
\includegraphics[width=\wxp\linewidth,height=\wxp\linewidth,keepaspectratio]{figures/appendix/57/1}  &
\includegraphics[width=\wxp\linewidth,height=\wxp\linewidth,keepaspectratio]{figures/appendix/57/2}  &
\includegraphics[width=\wxp\linewidth,height=\wxp\linewidth,keepaspectratio]{figures/appendix/57/3}  &
\includegraphics[width=\wxp\linewidth,height=\wxp\linewidth,keepaspectratio]{figures/appendix/57/4}  &
\includegraphics[width=\wxp\linewidth,height=\wxp\linewidth,keepaspectratio]{figures/appendix/58/0}  &
\includegraphics[width=\wxp\linewidth,height=\wxp\linewidth,keepaspectratio]{figures/appendix/58/1}  &
\includegraphics[width=\wxp\linewidth,height=\wxp\linewidth,keepaspectratio]{figures/appendix/58/2}  &
\includegraphics[width=\wxp\linewidth,height=\wxp\linewidth,keepaspectratio]{figures/appendix/58/3}  &
\includegraphics[width=\wxp\linewidth,height=\wxp\linewidth,keepaspectratio]{figures/appendix/58/4} \\ 

\includegraphics[width=\wxp\linewidth,height=\wxp\linewidth,keepaspectratio]{figures/appendix/57/0_pred}  &
\includegraphics[width=\wxp\linewidth,height=\wxp\linewidth,keepaspectratio]{figures/appendix/57/1_pred}  &
\includegraphics[width=\wxp\linewidth,height=\wxp\linewidth,keepaspectratio]{figures/appendix/57/2_pred}  &
\includegraphics[width=\wxp\linewidth,height=\wxp\linewidth,keepaspectratio]{figures/appendix/57/3_pred}  &
\includegraphics[width=\wxp\linewidth,height=\wxp\linewidth,keepaspectratio]{figures/appendix/57/4_pred}  &
\includegraphics[width=\wxp\linewidth,height=\wxp\linewidth,keepaspectratio]{figures/appendix/58/0_pred}  &
\includegraphics[width=\wxp\linewidth,height=\wxp\linewidth,keepaspectratio]{figures/appendix/58/1_pred}  &
\includegraphics[width=\wxp\linewidth,height=\wxp\linewidth,keepaspectratio]{figures/appendix/58/2_pred}  &
\includegraphics[width=\wxp\linewidth,height=\wxp\linewidth,keepaspectratio]{figures/appendix/58/3_pred}  &
\includegraphics[width=\wxp\linewidth,height=\wxp\linewidth,keepaspectratio]{figures/appendix/58/4_pred} \\ 

\includegraphics[width=\wxp\linewidth,height=\wxp\linewidth,keepaspectratio]{figures/appendix/76/0}  &
\includegraphics[width=\wxp\linewidth,height=\wxp\linewidth,keepaspectratio]{figures/appendix/76/1}  &
\includegraphics[width=\wxp\linewidth,height=\wxp\linewidth,keepaspectratio]{figures/appendix/76/2}  &
\includegraphics[width=\wxp\linewidth,height=\wxp\linewidth,keepaspectratio]{figures/appendix/76/3}  &
\includegraphics[width=\wxp\linewidth,height=\wxp\linewidth,keepaspectratio]{figures/appendix/76/4}  &
\includegraphics[width=\wxp\linewidth,height=\wxp\linewidth,keepaspectratio]{figures/appendix/81/0}  &
\includegraphics[width=\wxp\linewidth,height=\wxp\linewidth,keepaspectratio]{figures/appendix/81/1}  &
\includegraphics[width=\wxp\linewidth,height=\wxp\linewidth,keepaspectratio]{figures/appendix/81/2}  &
\includegraphics[width=\wxp\linewidth,height=\wxp\linewidth,keepaspectratio]{figures/appendix/81/3}  &
\includegraphics[width=\wxp\linewidth,height=\wxp\linewidth,keepaspectratio]{figures/appendix/81/4} \\ 

\includegraphics[width=\wxp\linewidth,height=\wxp\linewidth,keepaspectratio]{figures/appendix/76/0_pred}  &
\includegraphics[width=\wxp\linewidth,height=\wxp\linewidth,keepaspectratio]{figures/appendix/76/1_pred}  &
\includegraphics[width=\wxp\linewidth,height=\wxp\linewidth,keepaspectratio]{figures/appendix/76/2_pred}  &
\includegraphics[width=\wxp\linewidth,height=\wxp\linewidth,keepaspectratio]{figures/appendix/76/3_pred}  &
\includegraphics[width=\wxp\linewidth,height=\wxp\linewidth,keepaspectratio]{figures/appendix/76/4_pred}  &
\includegraphics[width=\wxp\linewidth,height=\wxp\linewidth,keepaspectratio]{figures/appendix/81/0_pred}  &
\includegraphics[width=\wxp\linewidth,height=\wxp\linewidth,keepaspectratio]{figures/appendix/81/1_pred}  &
\includegraphics[width=\wxp\linewidth,height=\wxp\linewidth,keepaspectratio]{figures/appendix/81/2_pred}  &
\includegraphics[width=\wxp\linewidth,height=\wxp\linewidth,keepaspectratio]{figures/appendix/81/3_pred}  &
\includegraphics[width=\wxp\linewidth,height=\wxp\linewidth,keepaspectratio]{figures/appendix/81/4_pred} \\ 

\includegraphics[width=\wxp\linewidth,height=\wxp\linewidth,keepaspectratio]{figures/appendix/96/0}  &
\includegraphics[width=\wxp\linewidth,height=\wxp\linewidth,keepaspectratio]{figures/appendix/96/1}  &
\includegraphics[width=\wxp\linewidth,height=\wxp\linewidth,keepaspectratio]{figures/appendix/96/2}  &
\includegraphics[width=\wxp\linewidth,height=\wxp\linewidth,keepaspectratio]{figures/appendix/96/3}  &
\includegraphics[width=\wxp\linewidth,height=\wxp\linewidth,keepaspectratio]{figures/appendix/96/4}  &
\includegraphics[width=\wxp\linewidth,height=\wxp\linewidth,keepaspectratio]{figures/appendix/111/0}  &
\includegraphics[width=\wxp\linewidth,height=\wxp\linewidth,keepaspectratio]{figures/appendix/111/1}  &
\includegraphics[width=\wxp\linewidth,height=\wxp\linewidth,keepaspectratio]{figures/appendix/111/2}  &
\includegraphics[width=\wxp\linewidth,height=\wxp\linewidth,keepaspectratio]{figures/appendix/111/3}  &
\includegraphics[width=\wxp\linewidth,height=\wxp\linewidth,keepaspectratio]{figures/appendix/111/4} \\ 

\includegraphics[width=\wxp\linewidth,height=\wxp\linewidth,keepaspectratio]{figures/appendix/96/0_pred}  &
\includegraphics[width=\wxp\linewidth,height=\wxp\linewidth,keepaspectratio]{figures/appendix/96/1_pred}  &
\includegraphics[width=\wxp\linewidth,height=\wxp\linewidth,keepaspectratio]{figures/appendix/96/2_pred}  &
\includegraphics[width=\wxp\linewidth,height=\wxp\linewidth,keepaspectratio]{figures/appendix/96/3_pred}  &
\includegraphics[width=\wxp\linewidth,height=\wxp\linewidth,keepaspectratio]{figures/appendix/96/4_pred}  &
\includegraphics[width=\wxp\linewidth,height=\wxp\linewidth,keepaspectratio]{figures/appendix/111/0_pred}  &
\includegraphics[width=\wxp\linewidth,height=\wxp\linewidth,keepaspectratio]{figures/appendix/111/1_pred}  &
\includegraphics[width=\wxp\linewidth,height=\wxp\linewidth,keepaspectratio]{figures/appendix/111/2_pred}  &
\includegraphics[width=\wxp\linewidth,height=\wxp\linewidth,keepaspectratio]{figures/appendix/111/3_pred}  &
\includegraphics[width=\wxp\linewidth,height=\wxp\linewidth,keepaspectratio]{figures/appendix/111/4_pred} \\ 

\includegraphics[width=\wxp\linewidth,height=\wxp\linewidth,keepaspectratio]{figures/appendix/127/0}  &
\includegraphics[width=\wxp\linewidth,height=\wxp\linewidth,keepaspectratio]{figures/appendix/127/1}  &
\includegraphics[width=\wxp\linewidth,height=\wxp\linewidth,keepaspectratio]{figures/appendix/127/2}  &
\includegraphics[width=\wxp\linewidth,height=\wxp\linewidth,keepaspectratio]{figures/appendix/127/3}  &
\includegraphics[width=\wxp\linewidth,height=\wxp\linewidth,keepaspectratio]{figures/appendix/127/4}  &
\includegraphics[width=\wxp\linewidth,height=\wxp\linewidth,keepaspectratio]{figures/appendix/160/0}  &
\includegraphics[width=\wxp\linewidth,height=\wxp\linewidth,keepaspectratio]{figures/appendix/160/1}  &
\includegraphics[width=\wxp\linewidth,height=\wxp\linewidth,keepaspectratio]{figures/appendix/160/2}  &
\includegraphics[width=\wxp\linewidth,height=\wxp\linewidth,keepaspectratio]{figures/appendix/160/3}  &
\includegraphics[width=\wxp\linewidth,height=\wxp\linewidth,keepaspectratio]{figures/appendix/160/4} \\ 

\includegraphics[width=\wxp\linewidth,height=\wxp\linewidth,keepaspectratio]{figures/appendix/127/0_pred}  &
\includegraphics[width=\wxp\linewidth,height=\wxp\linewidth,keepaspectratio]{figures/appendix/127/1_pred}  &
\includegraphics[width=\wxp\linewidth,height=\wxp\linewidth,keepaspectratio]{figures/appendix/127/2_pred}  &
\includegraphics[width=\wxp\linewidth,height=\wxp\linewidth,keepaspectratio]{figures/appendix/127/3_pred}  &
\includegraphics[width=\wxp\linewidth,height=\wxp\linewidth,keepaspectratio]{figures/appendix/127/4_pred}  &
\includegraphics[width=\wxp\linewidth,height=\wxp\linewidth,keepaspectratio]{figures/appendix/160/0_pred}  &
\includegraphics[width=\wxp\linewidth,height=\wxp\linewidth,keepaspectratio]{figures/appendix/160/1_pred}  &
\includegraphics[width=\wxp\linewidth,height=\wxp\linewidth,keepaspectratio]{figures/appendix/160/2_pred}  &
\includegraphics[width=\wxp\linewidth,height=\wxp\linewidth,keepaspectratio]{figures/appendix/160/3_pred}  &
\includegraphics[width=\wxp\linewidth,height=\wxp\linewidth,keepaspectratio]{figures/appendix/160/4_pred} \\ 

\includegraphics[width=\wxp\linewidth,height=\wxp\linewidth,keepaspectratio]{figures/appendix/196/0}  &
\includegraphics[width=\wxp\linewidth,height=\wxp\linewidth,keepaspectratio]{figures/appendix/196/1}  &
\includegraphics[width=\wxp\linewidth,height=\wxp\linewidth,keepaspectratio]{figures/appendix/196/2}  &
\includegraphics[width=\wxp\linewidth,height=\wxp\linewidth,keepaspectratio]{figures/appendix/196/3}  &
\includegraphics[width=\wxp\linewidth,height=\wxp\linewidth,keepaspectratio]{figures/appendix/196/4}  &
\includegraphics[width=\wxp\linewidth,height=\wxp\linewidth,keepaspectratio]{figures/appendix/214/0}  &
\includegraphics[width=\wxp\linewidth,height=\wxp\linewidth,keepaspectratio]{figures/appendix/214/1}  &
\includegraphics[width=\wxp\linewidth,height=\wxp\linewidth,keepaspectratio]{figures/appendix/214/2}  &
\includegraphics[width=\wxp\linewidth,height=\wxp\linewidth,keepaspectratio]{figures/appendix/214/3}  &
\includegraphics[width=\wxp\linewidth,height=\wxp\linewidth,keepaspectratio]{figures/appendix/214/4} \\ 

\includegraphics[width=\wxp\linewidth,height=\wxp\linewidth,keepaspectratio]{figures/appendix/196/0_pred}  &
\includegraphics[width=\wxp\linewidth,height=\wxp\linewidth,keepaspectratio]{figures/appendix/196/1_pred}  &
\includegraphics[width=\wxp\linewidth,height=\wxp\linewidth,keepaspectratio]{figures/appendix/196/2_pred}  &
\includegraphics[width=\wxp\linewidth,height=\wxp\linewidth,keepaspectratio]{figures/appendix/196/3_pred}  &
\includegraphics[width=\wxp\linewidth,height=\wxp\linewidth,keepaspectratio]{figures/appendix/196/4_pred}  &
\includegraphics[width=\wxp\linewidth,height=\wxp\linewidth,keepaspectratio]{figures/appendix/214/0_pred}  &
\includegraphics[width=\wxp\linewidth,height=\wxp\linewidth,keepaspectratio]{figures/appendix/214/1_pred}  &
\includegraphics[width=\wxp\linewidth,height=\wxp\linewidth,keepaspectratio]{figures/appendix/214/2_pred}  &
\includegraphics[width=\wxp\linewidth,height=\wxp\linewidth,keepaspectratio]{figures/appendix/214/3_pred}  &
\includegraphics[width=\wxp\linewidth,height=\wxp\linewidth,keepaspectratio]{figures/appendix/214/4_pred} \\ 

\includegraphics[width=\wxp\linewidth,height=\wxp\linewidth,keepaspectratio]{figures/appendix/252/0}  &
\includegraphics[width=\wxp\linewidth,height=\wxp\linewidth,keepaspectratio]{figures/appendix/252/1}  &
\includegraphics[width=\wxp\linewidth,height=\wxp\linewidth,keepaspectratio]{figures/appendix/252/2}  &
\includegraphics[width=\wxp\linewidth,height=\wxp\linewidth,keepaspectratio]{figures/appendix/252/3}  &
\includegraphics[width=\wxp\linewidth,height=\wxp\linewidth,keepaspectratio]{figures/appendix/252/4}  &
\includegraphics[width=\wxp\linewidth,height=\wxp\linewidth,keepaspectratio]{figures/appendix/264/0}  &
\includegraphics[width=\wxp\linewidth,height=\wxp\linewidth,keepaspectratio]{figures/appendix/264/1}  &
\includegraphics[width=\wxp\linewidth,height=\wxp\linewidth,keepaspectratio]{figures/appendix/264/2}  &
\includegraphics[width=\wxp\linewidth,height=\wxp\linewidth,keepaspectratio]{figures/appendix/264/3}  &
\includegraphics[width=\wxp\linewidth,height=\wxp\linewidth,keepaspectratio]{figures/appendix/264/4} \\ 

\includegraphics[width=\wxp\linewidth,height=\wxp\linewidth,keepaspectratio]{figures/appendix/252/0_pred}  &
\includegraphics[width=\wxp\linewidth,height=\wxp\linewidth,keepaspectratio]{figures/appendix/252/1_pred}  &
\includegraphics[width=\wxp\linewidth,height=\wxp\linewidth,keepaspectratio]{figures/appendix/252/2_pred}  &
\includegraphics[width=\wxp\linewidth,height=\wxp\linewidth,keepaspectratio]{figures/appendix/252/3_pred}  &
\includegraphics[width=\wxp\linewidth,height=\wxp\linewidth,keepaspectratio]{figures/appendix/252/4_pred}  &
\includegraphics[width=\wxp\linewidth,height=\wxp\linewidth,keepaspectratio]{figures/appendix/264/0_pred}  &
\includegraphics[width=\wxp\linewidth,height=\wxp\linewidth,keepaspectratio]{figures/appendix/264/1_pred}  &
\includegraphics[width=\wxp\linewidth,height=\wxp\linewidth,keepaspectratio]{figures/appendix/264/2_pred}  &
\includegraphics[width=\wxp\linewidth,height=\wxp\linewidth,keepaspectratio]{figures/appendix/264/3_pred}  &
\includegraphics[width=\wxp\linewidth,height=\wxp\linewidth,keepaspectratio]{figures/appendix/264/4_pred} \\ 
\end{tabular}
\caption{Reconstructions of the ShapeNet testing set.}
\end{figure}

\newpage
\section{Network Input Gate Analysis}
\label{sec:analysis}

\begin{figure}[htp!]
\centering
\scriptsize
\begin{tabular}{MM}
  (a) & \includegraphics[width=0.9\linewidth]{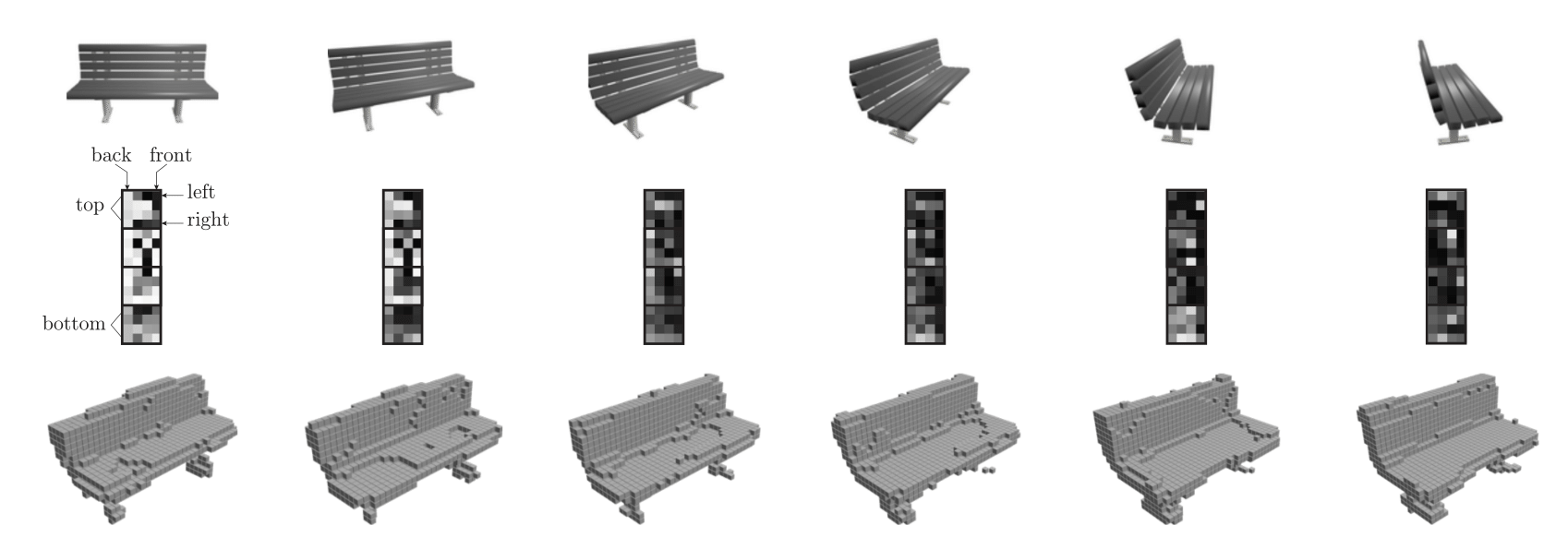} \\
  (b) & \includegraphics[width=0.9\linewidth]{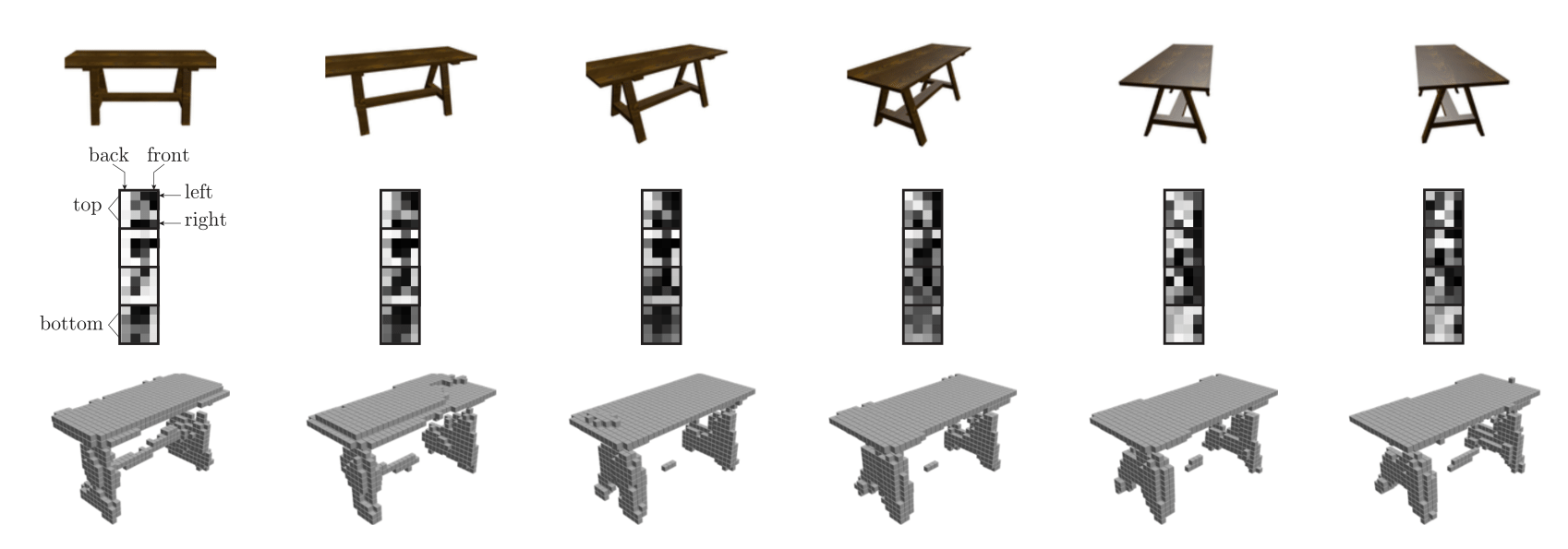} \\
  (c) & \includegraphics[width=0.9\linewidth]{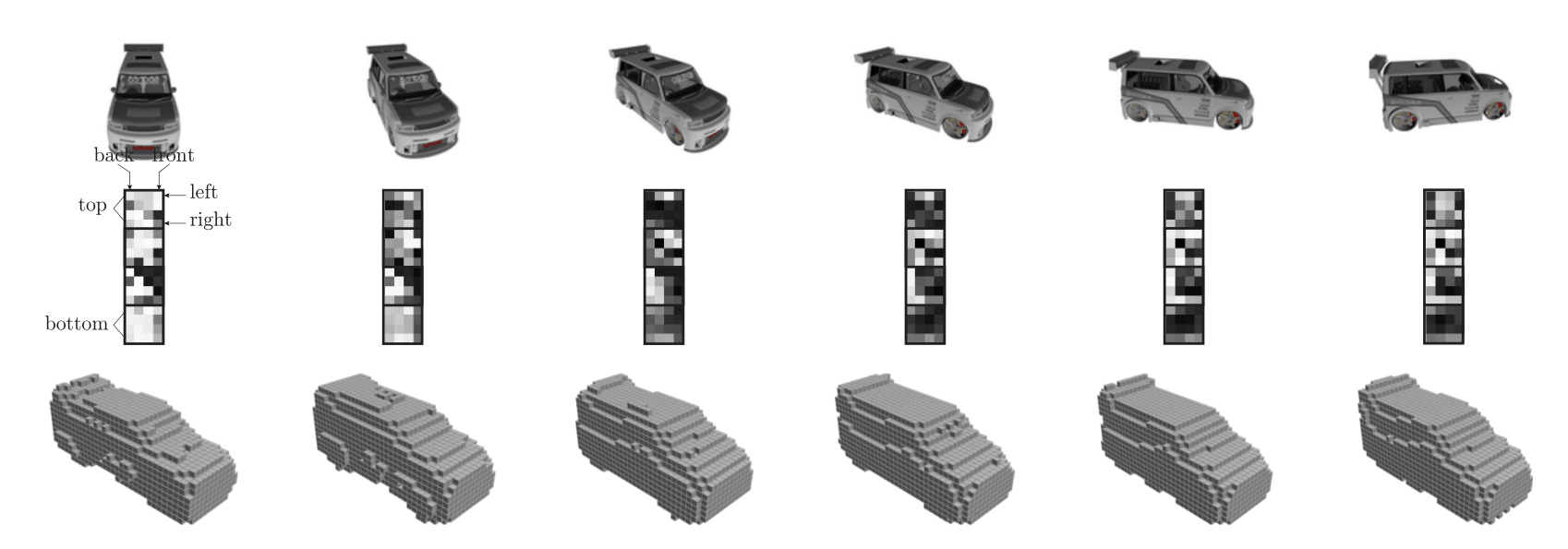} \\
  (d) & \includegraphics[width=0.9\linewidth]{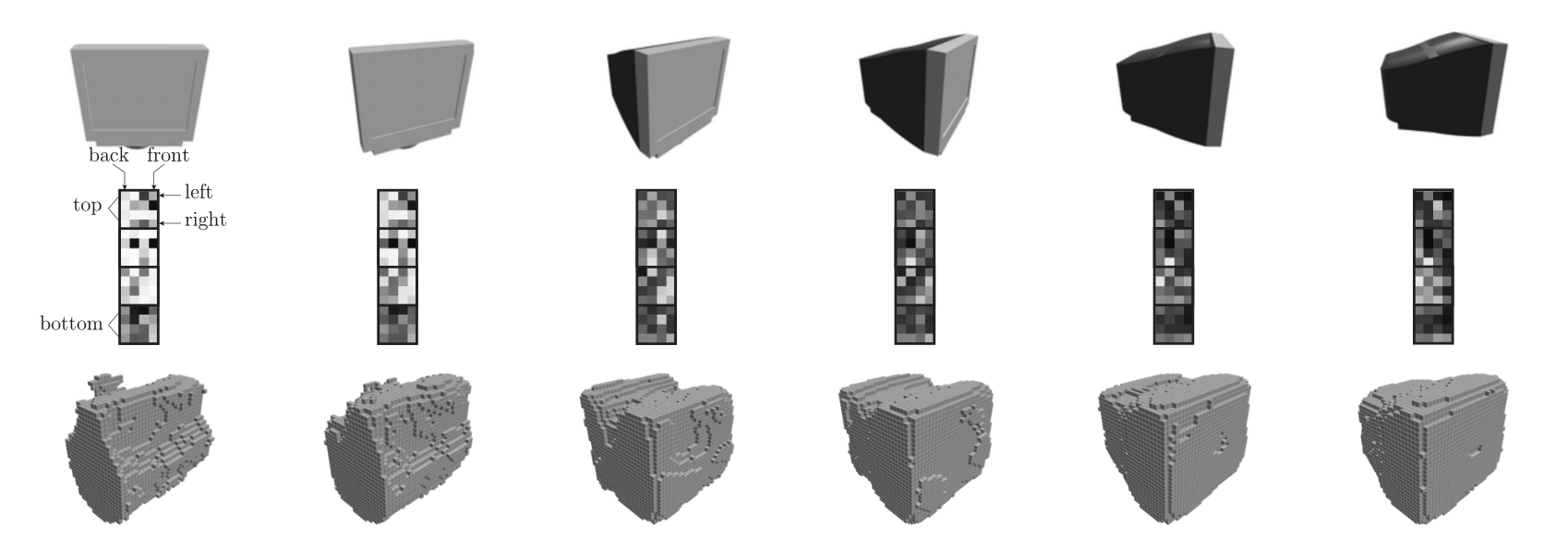}
\end{tabular}
\caption{Input gate channel \#48 activations as the network observes more images. Top row: input images $x_t$, Center row: Input gate activations corresponding to the input image. White indicates strong activation (gate open), and black indicates no activation (gate closed). Bottom row: corresponding reconstructions.}
\label{fig:inputgate}
\end{figure}

In this section, we visualize the network input gate activations to observe how
the input gates react as new viewpoints become available. We used the
[Res3D-GRU-3] network and tested on ShapeNet images taken sequentially from the front
viewpoint to the side viewpoint for this controlled experiment. We set the
viewpoints to be the same for all experiments to make the analysis easier, and
the activations are shown in Fig.~\ref{fig:inputgate}. First, we fed in the images
sequentially  (Fig.~\ref{fig:inputgate} top row) and selected a specific channel
of the input gate activation to analyze, giving us a $4\times4\times 4$ tensor.
To visualize this tensor, we extracted four $4\times 4$ grids, with the $i$th
grid corresponding to the $i$th slice of the $4\times4\times 4$ tensor in the
3rd dimension. Lastly, the grids are concatenated to create a $16 \times 4$ grid
in a manner such that the top grid corresponds to activations at the top of the object
and the bottom grid corresponds to activations at the bottom of the object.


For this particular example, we show the input gate activations
from channel \#48. The input gates open up strongly at first to make adjustments to the
predictions but soon close down as the predictions get more accurate. However,
when a part of the prediction mismatches the observation, the corresponding input gates
open up to update the prediction.


%% file: abstract.tex
\begin{abstract}

Inspired by the recent success of methods that employ shape priors to achieve robust 3D reconstructions, we propose a novel recurrent neural network architecture that we call the 3D Recurrent Reconstruction Neural Network (3D-R2N2). The network learns a mapping from images of objects to their underlying 3D shapes from a large collection of synthetic data~\cite{shapenet}. Our network takes in one or more images of an object instance
from arbitrary viewpoints and outputs a reconstruction of the object in the form of a 3D  occupancy  grid. Unlike most of the previous works, our network does not require any image annotations or object class labels for training or testing.
Our extensive experimental analysis shows that our  reconstruction framework i) outperforms the state-of-the-art methods for single view reconstruction, and ii) enables the 3D reconstruction  of  objects in situations when traditional SFM/SLAM methods fail (because of lack of texture and/or wide baseline).

\keywords{multi-view, reconstruction, recurrent neural network}
\end{abstract}

%% file: intro.tex
\section{Introduction}




Rapid and automatic 3D object prototyping has become a game-changing innovation in many applications related to e-commerce, visualization, and architecture, to name a few. This trend has been boosted now that 3D printing is a democratized technology and 3D acquisition methods are accurate and efficient~\cite{choi2016large}.
Moreover, the trend is also coupled with the diffusion of large scale repositories of 3D object models such as ShapeNet~\cite{shapenet}.

\begin{figure}[!htp]
  \centering
  \includegraphics[width=0.9\textwidth]{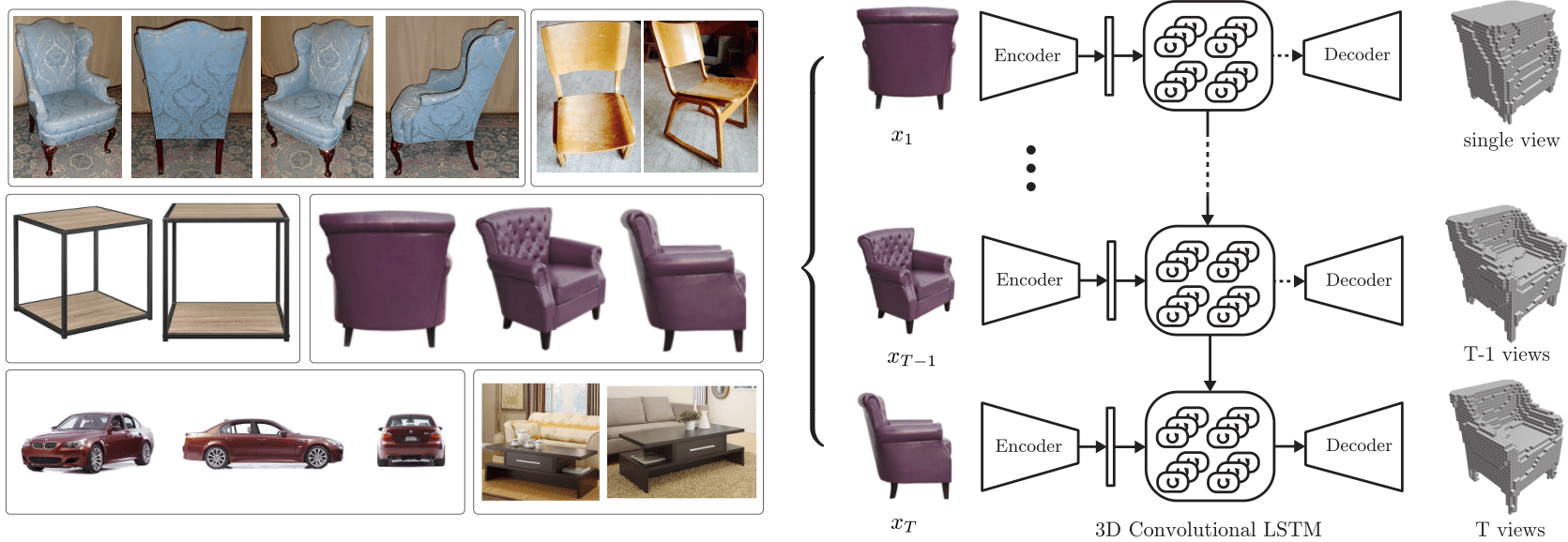}
  \begin{tabular}{cc}
    (a) Images of objects we wish to reconstruct & (b) Overview of the network
  \end{tabular}
  \caption{ (a) Some sample images of the objects we wish to reconstruct - notice that views are separated by a large baseline and objects' appearance shows little texture and/or are non-lambertian. (b) An overview of our proposed \textbf{3D-R2N2}: The network takes a sequence of images (or just one image) from arbitrary (uncalibrated) viewpoints as input (in this example, 3 views of the armchair) and generates voxelized 3D reconstruction as an output. The reconstruction is incrementally refined as the network sees more views of the object. }
  \label{fig:overview}
  \vspace{\figurevspace}
\end{figure}

Most of the state-of-the-art methods for 3D object reconstruction, however, are subject to a number of restrictions. Some restrictions are that: i) objects must be observed from a dense number of views; or equivalently, views must have a relatively small baseline. This is an issue when users wish to reconstruct the object from just a handful of views or ideally just one view (see Fig.~\ref{fig:overview}(a)); ii) objects' appearances (or their reflectance functions) are expected to be Lambertian (i.e. non-reflective) and the albedos are supposed be non-uniform (i.e., rich of non-homogeneous textures). 

These restrictions stem from a number of key technical assumptions. One typical assumption is that features can be matched across views \cite{Fitzgibbon1998sfm,lhuillier2005sfm,agarwal2009rome,engel2014lsd} as hypothesized by the majority of the methods based on SFM or SLAM~\cite{haming2010sfmsurvey,fuentes2015slamsurvey}. It has been demonstrated (for instance see~\cite{SIFT}) that if the viewpoints are separated by a large baseline, establishing (traditional) feature correspondences is extremely problematic due to local appearance changes
or self-occlusions. Moreover, lack of texture on objects and specular reflections also make the feature matching problem very difficult~\cite{bhat1998specular,saponaro2014sfmtextureless}.  

In order to circumvent issues related to large baselines or non-Lambertian surfaces, 3D volumetric reconstruction methods such as space carving~\cite{seitz1999voxelcoloring,kutulako2000spacecarving,slabaugh2004voxelcoloring,anwar2006voxelcoloring} and their probabilistic extensions~\cite{broadhurst2001probabilistic} have become popular. These methods, however, assume that the objects are accurately segmented from the background or that the cameras are calibrated, which is not the case in many applications. 

A different philosophy is to assume that prior knowledge about the object appearance and shape is available. The benefit of using priors is that the ensuing reconstruction method is less reliant on finding accurate feature correspondences across views. Thus, shape prior-based methods can work with fewer images and with fewer assumptions on the object reflectance function as shown in~\cite{dame,bao}. The shape priors are typically encoded in the form of simple 3D primitives as demonstrated by early pioneering works~\cite{lawrence1963single,nevatia1977single} or learned from rich repositories of 3D CAD models~\cite{zia2013detailed,rock2015completing,choy}, whereby the concept of fitting 3D models to images of faces was explored to a much larger extent~\cite{blanz2003face,matthews2007face,kemelmacher2011face}. Sophisticated mathematical formulations have also been introduced to adapt 3D shape models to observations with different degrees of supervision~\cite{prisacariu2012accv} and different regularization strategies~\cite{sandhu2011nonrigid}.

This paper is in the same spirit as the methods discussed above, but with a key difference. Instead of trying to match a suitable 3D shape prior to the observation of the object and possibly adapt to it, we use deep convolutional neural networks to learn a mapping from observations to their underlying 3D shapes of objects from a large collection of training data. Inspired by early works that used machine learning to learn a 2D-to-3D mapping for scene understanding~\cite{saxena_make3d,hoiem2005popup}, data driven approaches have been recently proposed to solve the daunting problem of recovering the shape of an object from just a single image~\cite{vicente,kar2015category} for a given number of object categories. In our approach, however, we leverage for the first time the ability of deep neural networks to automatically learn, in a mere end-to-end fashion, the appropriate intermediate representations from data to recover approximated 3D object reconstructions from as few as a single image with minimal supervision. 

Inspired by the success of Long Short-Term Memory (LSTM)~\cite{lstm} networks~\cite{sundermeyer2012lstm,sutskever2014lstm} as well as recent progress in single-view 3D reconstruction using Convolutional Neural Networks
~\cite{eigen_depth,liu_2015}, we propose a novel architecture that we call the 3D Recurrent Reconstruction Neural Network (3D-R2N2). The network takes in one or more images of an object instance from different viewpoints and outputs a reconstruction of the object in the form of a 3D occupancy grid, as illustrated in Fig. \ref{fig:overview}(b). Note that in both training and testing, our network does not require any object class labels or image annotations (i.e., no segmentations, keypoints, viewpoint labels, or class labels are needed).


One of the key attributes of the 3D-R2N2 is that it can selectively update hidden representations by controlling \emph{input} gates and \emph{forget} gates. In training, this mechanism allows the network to adaptively and consistently learn a suitable 3D representation of an object as (potentially conflicting) information from different viewpoints becomes available (see Fig. \ref{fig:overview}). 


The main contributions of this paper are summarized as follows:

\vspace{-0.2cm}
\begin{itemize}
\item We propose an extension of the standard LSTM framework that we call the 3D Recurrent Reconstruction Neural Network which is suitable for accommodating multi-view image feeds in a principled manner.
\item We unify single- and multi-view 3D reconstruction in a single framework.
\item Our approach requires minimal supervision in training and testing (just bounding boxes, but no segmentation, keypoints, viewpoint labels, camera calibration, or class labels are needed).
\item Our extensive experimental analysis shows that our reconstruction framework outperforms the state-of-the-art method for single-view reconstruction~\cite{kar2015category}. 
\item Our network enables the 3D reconstruction of objects in situations when traditional SFM/SLAM methods fail (because of lack of texture or wide baselines).
\end{itemize}
\vspace{-0.2cm}

An overview of our reconstruction network is shown in Fig.~\ref{fig:overview}(b). The rest of this paper is organized as follows. In Section~\ref{sec:rnn}, we give a brief overview of LSTM and GRU networks. In Section~\ref{sec:network}, we introduce the 3D Recurrent Reconstruction Neural Network architecture. In Section~\ref{sec:implementation}, we discuss how we generate training data and give details of the training process. Finally, we present test results of our approach on various datasets including PASCAL 3D and ShapeNet in Section~\ref{sec:experiments}.

%% file: rnn.tex
\section{Recurrent Neural Network}
\label{sec:rnn}

In this section we provide a brief overview of Long Short-Term Memory (LSTM) networks and a variation of the LSTM called Gated Recurrent Units (GRU). 

\textbf{Long Short-Term Memory Unit.} One of the most successful implementations of the hidden states of an RNN is the Long Short Term Memory (LSTM) unit~\cite{lstm}. An LSTM unit explicitly controls the flow from input to output, allowing the network to overcome the vanishing gradient problem~\cite{lstm,rnn_difficult}. Specifically, an LSTM unit consists of four components: memory units (a memory cell and a hidden state), and three gates which control the flow of information from the input to the hidden state (\textit{input gate}), from the hidden state to the output (\textit{output gate}), and from the previous hidden state to the current hidden state (\textit{forget gate}). More formally, at time step $t$ when a new input $x_t$ is received, the operation of an LSTM unit can be expressed as:

\vspace{\alignvspace}
\begin{align}
  i_t & = \sigma(W_{i} x_t + U_{i} h_{t-1} + b_i) \\
  f_t & = \sigma(W_{f} x_t + U_{f} h_{t-1} + b_f) \\  
  o_t & = \sigma(W_{o} x_t + U_{o} h_{t-1} + b_o) \\
  s_t & = f_t \odot s_{t-1} + i_t \odot \tanh(W_{s} x_t + U_{s} h_{t-1} + b_s) \\
  h_t & = o_t \odot \tanh(s_t)
\end{align}
\vspace{\alignvspace}

$i_t, f_t, o_t$ refer to the input gate, the output gate, and the forget gate, respectively. $s_t$ and $h_t$ refer to the memory cell and the hidden state, respectively. We use $\odot$ to denote element-wise multiplication and the subscript $t$ to refer to an activation at time $t$. $W_{(\cdot)}, U_{(\cdot)}$ are matrices that transform the current input $x_t$ and the previous hidden state $h_{t-1}$, respectively, and $b_{(\cdot)}$ represents the biases.

\textbf{Gated Recurrent Unit.} 
A variation of the LSTM unit is the Gated Recurrent Unit (GRU) proposed by Cho et al.~\cite{gru}. An advantage of the GRU is that there are fewer computations compared to the standard LSTM. In a GRU, an update gate controls both the input and forget gates. Another difference is that a \textit{reset gate} is applied before the nonlinear transformation. More formally,

\vspace{\alignvspace}
\begin{align}
  u_t & = \sigma(W_{u} \mathcal{T}x_t + U_{u} \ast h_{t-1} + b_f) \\
  r_t & = \sigma(W_{i} \mathcal{T}x_t + U_{i} \ast h_{t-1} + b_i) \\
  h_t & = (1 - u_t) \odot h_{t-1} + u_t \odot \tanh(W_h x_t + U_h (r_t \odot h_{t-1}) + b_h)
\end{align}
\vspace{\alignvspace}

$u_t, r_t, h_t$ represent the update gate, the reset gate, and the hidden state respectively. We follow the same notations as LSTM for matrices and biases.

%% file: network.tex
\section{3D Recurrent Reconstruction Neural Network}
\label{sec:network}

In this section, we introduce a novel architecture named the 3D Recurrent Reconstruction Network
(3D-R2N2), which builds upon the standard LSTM and GRU. The goal of the network is to perform both single- and multi-view 3D reconstructions. The main idea is to leverage the power of LSTM to retain previous observations and incrementally refine the output reconstruction as more observations become available. 


\begin{figure}[t]
\centering
\scriptsize
\includegraphics[width=0.99\linewidth]{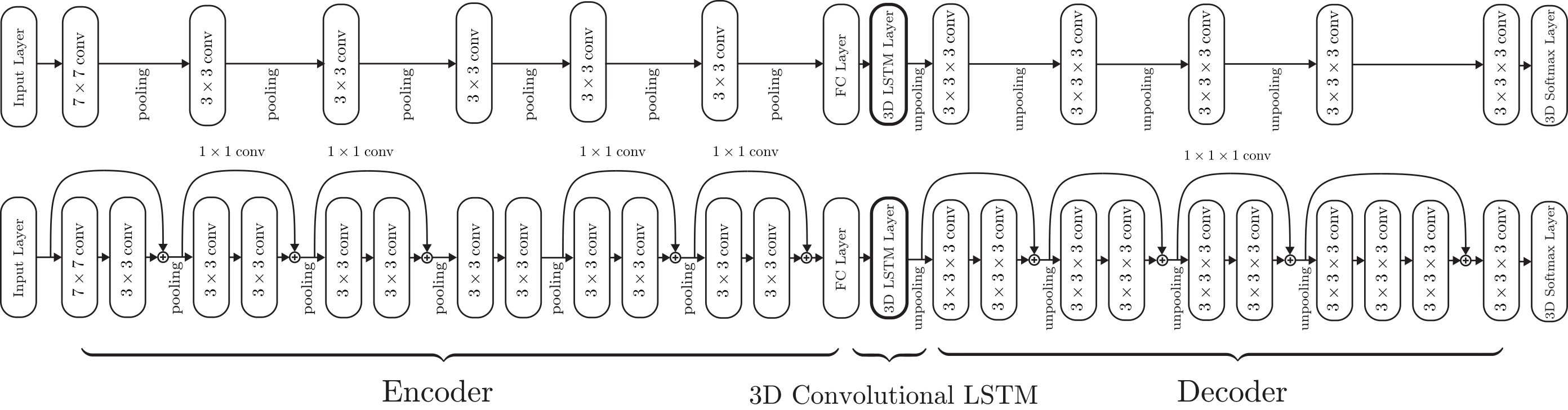}
\caption{Network architecture: Each 3D-R2N2 consists of an encoder, a recurrence unit and a decoder. After every convolution layer, we place a LeakyReLU nonlinearity. The encoder converts a $127 \times 127$ RGB image into a low-dimensional feature which is then fed into the 3D-LSTM. The decoder then takes the 3D-LSTM hidden states and transforms them to a final voxel occupancy map. After each convolution layer is a LeakyReLU. We use two versions of 3D-R2N2: (top) a shallow network and (bottom) a deep residual network~\cite{resnet}.}
\label{fig:network}
\vspace{\figurevspace}
\end{figure}

The network is made up of three components: a 2D Convolutional Neural Network (2D-CNN),
a novel architecture named 3D Convolutional LSTM (3D-LSTM), and a 3D Deconvolutional Neural Network (3D-DCNN) (see Fig.~\ref{fig:network}). Given one or more images of an object from arbitrary
viewpoints, the 2D-CNN first encodes each input image $x$ into low dimensional features $\mathcal{T}(x)$ (Section~\ref{sec:encoder}).
Then, given the encoded input, a set of newly proposed 3D Convolutional LSTM (3D-LSTM) units (Section~\ref{sec:recurrence})
either selectively update their cell states or retain the states by closing the input gate. 
Finally, the 3D-DCNN decodes the hidden states of the LSTM units and generates a 3D probabilistic voxel reconstruction (Section~\ref{sec:decoder}).

The main advantage of using an LSTM-based network comes from its ability to effectively
handle object self-occlusions when multiple views are fed to the network.
The network selectively updates the memory cells that correspond to the
visible parts of the object. If a subsequent view shows parts that were
previously self-occluded and mismatch the prediction, the network would update the LSTM states for
the previously occluded sections but retain the states of the other parts (Fig.~\ref{fig:network}).


\subsection{Encoder: 2D-CNN}
\label{sec:encoder}

We use CNNs to encode images into features.
We designed two different 2D-CNN encoders as shown in Fig.~\ref{fig:network}:
A standard feed-forward CNN and a deep residual variation of it.
The first network consists of standard convolution
layers, pooling layers, and leaky rectified 
linear units followed by a fully-connected layer. Motivated by a recent
study~\cite{resnet}, we also created a deep residual variation of the first network and report the performance of this variation in Section~\ref{sec:network_comparison}.
According to the study, adding residual connections between
standard convolution layers effectively improves
and speeds up the optimization process for very deep networks.
The deep residual variation of the encoder network
has identity mapping connections after every 2 convolution
layers except for the 4th pair. To match the number of channels after convolutions,
we use a $1 \times 1$ convolution for residual connections.
The encoder output is then flattened and passed to a fully connected layer which
compresses the output into a 1024 dimensional feature vector.


\subsection{Recurrence: 3D Convolutional LSTM}
\label{sec:recurrence}

\vspace{\figurevspace}
\begin{figure}[htp!]
\centering
\scriptsize
\begin{tabular}{MMM}
  \includegraphics[width=0.3\linewidth]{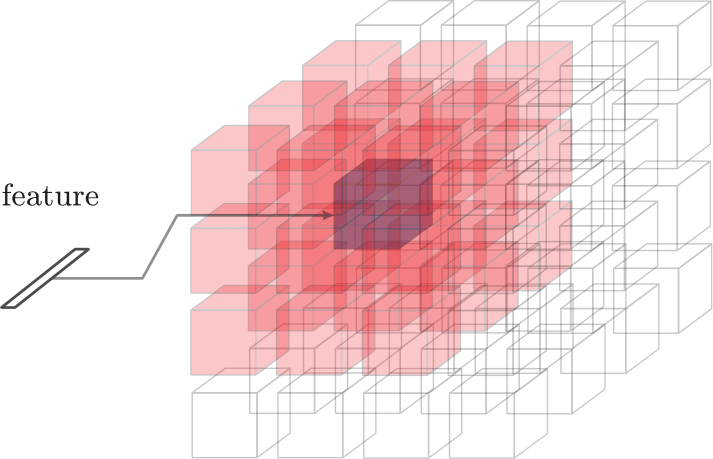} &
  \includegraphics[width=0.3\linewidth]{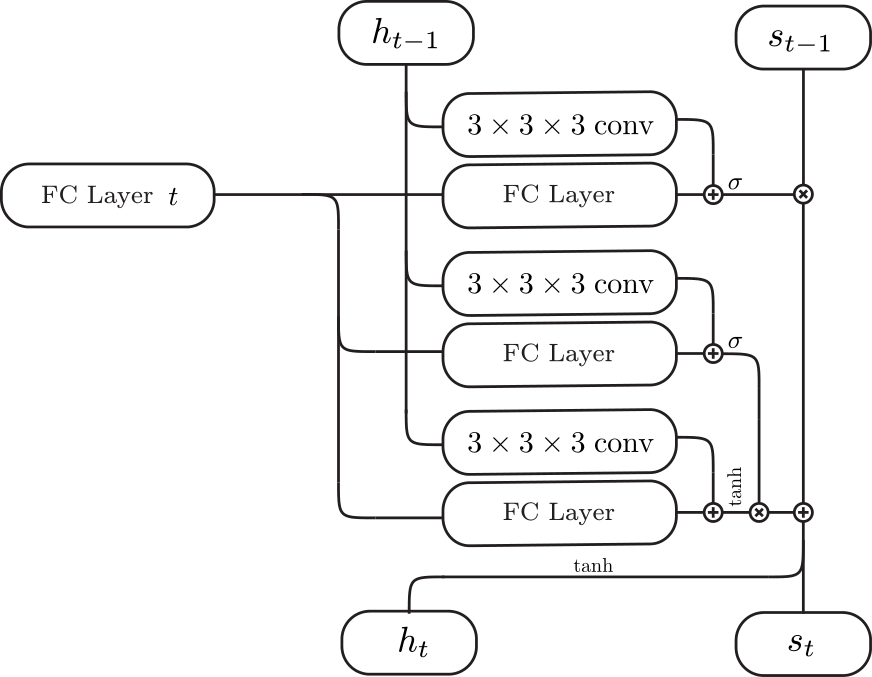} &
  \includegraphics[width=0.3\linewidth]{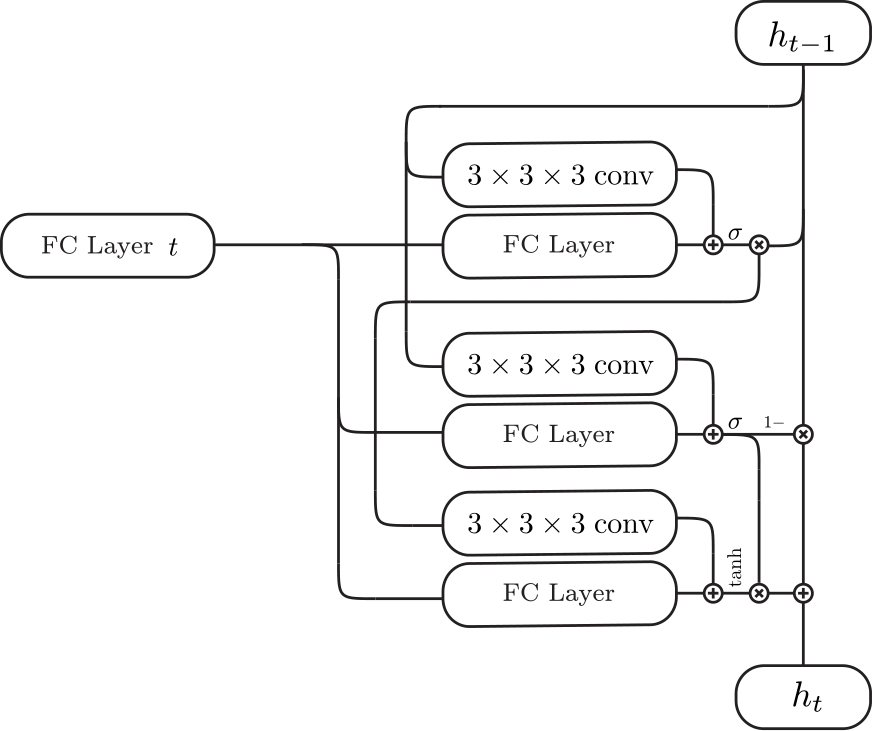} \\
  (a) inputs for each LSTM unit & (b) 3D Convolutional LSTMs & (c) 3D Convolutional GRUs
\end{tabular}
\caption{(a) At each time step, each unit (purple) in the 3D-LSTM receives the same feature vector from the encoder as well as the hidden states from its neighbors (red) by a $3\times 3\times 3$ convolution ($W_s \ast h_{t-1}$) as inputs. We propose two versions of 3D-LSTMs: (b) 3D-LSTMs
without output gates and (c) 3D Gated Recurrent Units (GRUs).}
\label{fig:lstm}
\vspace{\figurevspace}
\end{figure}

The core part of our 3D-R2N2 is a recurrence module that allows the network to retain what it has seen and to update the memory when it sees a new image. A naive approach would be to use a vanilla LSTM network. However, predicting such a large output space ($32\times 32\times 32$) would be a very difficult task without any regularization.
We propose a new architecture that we call 3D-Convolutional LSTM (3D-LSTM). The network is made up of a set of structured LSTM units with restricted connections. The 3D-LSTM units are spatially 
distributed in a 3D grid structure, with each unit responsible for 
reconstructing a particular part of the final output (see Fig.~\ref{fig:lstm}(a)). Inside the 3D grid, there are $N \times N \times N$ 3D-LSTM units where $N$ is the spatial resolution
of the 3D-LSTM grid. Each 3D-LSTM unit, indexed $(i, j, k)$, has an independent hidden state $h_{t,(i, j, k)} \in \mathbb{R}^{N_h}$.
Following the same notation as in Section~\ref{sec:rnn} but with  $f_t, i_t, s_t, h_t$ as 4D
tensors ($N \times N \times N$ vectors of size $N_h$), the equations governing the 3D-LSTM grid are

\vspace{\alignvspace}
\begin{align}
  f_t & = \sigma(W_{f} \mathcal{T}(x_t) + U_{f} \ast h_{t-1} + b_f) \\
  i_t & = \sigma(W_{i} \mathcal{T}(x_t) + U_{i} \ast h_{t-1} + b_i) \\
  s_t & = f_t \odot s_{t-1} + i_t \odot \tanh(W_{s} \mathcal{T}(x_t) + U_{s} \ast h_{t-1} + b_s) \\
  h_t & = \tanh(s_t)
\end{align}
\vspace{\alignvspace}

We denote the convolution operation as $\ast$. In our implementation, we use $N = 4$.
Unlike a standard LSTM, we do not have output gates since we only extract the
output at the end. By removing redundant output gates, we can reduce the number
of parameters.


Intuitively, this configuration forces a 3D-LSTM unit to handle the mismatch between
a particular region of the predicted reconstruction and the ground truth model such that
each unit learns to reconstruct one part of the voxel space instead of contributing to the
reconstruction of the entire space. This configuration also endows the network with a sense 
of locality so that it can selectively update its prediction about the 
previously occluded part of the object. We visualize such behavior in the appendix.

Moreover, a 3D Convolutional LSTM unit restricts the connections of
its hidden state to its spatial neighbors. 
For vanilla LSTMs, all elements in the hidden layer $h_{t-1}$ affect the current hidden state
$h_{t}$, whereas a spatially structured 3D Convolutional LSTM only allows its
hidden states $h_{t,(i, j, k)}$ to be affected by its neighboring 3D-LSTM units for all $i, j, \text{ and }k$. More specifically, the neighboring connections are defined by the convolution kernel size. For instance, if we use a $3 \times 3 \times 3$ kernel, an LSTM unit is only affected by its immediate neighbors.
This way, the units can share weights and the network can be further regularized.

In Section~\ref{sec:rnn}, we also described the Gated Recurrent Unit (GRU) as a variation of the LSTM unit.
We created a variation of the 3D-Convolutional LSTM using Gated Recurrent Unit (GRU). 
More formally, a GRU-based recurrence module can be expressed as

\vspace{\alignvspace}
\begin{align}
  u_t & = \sigma(W_{fx} \mathcal{T}(x_t) + U_{f} \ast h_{t-1} + b_f) \\
  r_t & = \sigma(W_{ix} \mathcal{T}(x_t) + U_{i} \ast h_{t-1} + b_i) \\
  h_t & = (1 - u_t) \odot h_{t-1} + u_t \odot \tanh(W_{h} \mathcal{T}(x_t) + U_{h} \ast (r_t \odot h_{t-1}) + b_h)
\end{align}
\vspace{\alignvspace}

\subsection{Decoder: 3D Deconvolutional Neural Network}
\label{sec:decoder}


After receiving an input image sequence $x_1, x_2, \cdots, x_T$, the 3D-LSTM passes the hidden state $h_T$
to a decoder, which increases the hidden state resolution by applying 3D convolutions, non-linearities, and 3D unpooling~\cite{unpooling} until it reaches the target output resolution.

As with the encoders, we propose a simple decoder network with 5 convolutions and a deep
residual version with 4 residual connections followed by a final convolution.
After the last layer where the activation reaches the target output resolution, 
we convert the final activation $\mathcal{V} \in \mathbb{R}^{N_{vox} \times
N_{vox} \times N_{vox} \times 2}$ to the occupancy probability $p_{(i, j, k)}$ of the voxel cell at $(i, j, k)$ using voxel-wise softmax.

\subsection{Loss: 3D Voxel-wise Softmax}
\label{sec:loss}

The loss function of the network is defined as the sum of voxel-wise cross-entropy. Let the
final output at each voxel $(i, j, k)$ be Bernoulli distributions $[1 -
p_{(i, j, k)}, p_{(i, j, k)}]$, where the dependency on input $\mathcal{X} = \{x_t\}_{t \in \{1,\ldots,T\}}$ is omitted, and let the corresponding 
ground truth occupancy be $y_{(i, j, k)} \in \{0, 1\}$, then

\vspace{\alignvspace}
\begin{align}
  L(\mathcal{X}, y) & = \sum_{i,j,k} y_{(i, j, k)} \log (p_{(i, j, k)}) + (1 - y_{(i, j, k)}) \log(1 - p_{(i, j, k)})
\end{align}

%% file: implementation.tex
\section{Implementation}
\label{sec:implementation}

\textbf{Data augmentation:} In training, we used 3D CAD models for generating input images and ground truth
voxel occupancy maps. We first rendered the CAD models with a transparent background
and then augmented the input images with random crops from the PASCAL VOC 2012 dataset~\cite{everingham2011pascal}.
Also, we tinted the color of the models and randomly translated the images. Note that all viewpoints were sampled randomly.

\noindent\textbf{Training:} In training the network, we used variable length inputs ranging from one
image to an arbitrary number of images. More specifically, the input length (number of views) for each training example within a single mini-batch
was kept constant, but the input length of training examples across \emph{different} mini-batches varied randomly. 
This enabled the network to perform both single- and multi-view reconstruction. During training, 
we computed the loss only at the end of an input sequence in order to save both computational
power and memory. On the other hand, during test time we could access the intermediate reconstructions at each time step
by extracting the hidden states of the LSTM units. 


\noindent\textbf{Network:}
The input image size was set to $127 \times 127$. The output voxelized reconstruction was of size $32 \times 32 \times 32$. The networks used in the experiments were trained for $60,000$ iterations with a batch size of $36$ except for [Res3D-GRU-3] (See Table~\ref{table:network_comparison}), which needed a batch size of $24$ to fit in an NVIDIA Titan X GPU. For the LeakyReLU layers, the slope of the leak was set to 0.1 throughout the network. For deconvolution, we followed the unpooling scheme presented in~\cite{unpooling}. We used Theano~\cite{theano} to implement our network and used Adam~\cite{adam}
for the SGD update rule.



%% file: experiments.tex
\section{Experiments}
\label{sec:experiments}

In this section, we validate and demonstrate the capability of our approach with several experiments using the datasets described in Section~\ref{sec:dataset}. First, we show the results of different variations of the 3D-R2N2 (Section~\ref{sec:network_comparison}). 
Next, we compare the performance of our network on the PASCAL 3D~\cite{xiang2014beyond} dataset with that of a state-of-the-art method by Kar et al.~\cite{kar2015category} for single-view real-world image reconstruction (Section~\ref{sec:pascal}). Then we show the network's ability to perform multi-view reconstruction on the ShapeNet dataset~\cite{shapenet} and the Online Products dataset~\cite{ebay} (Section~\ref{sec:multiview_shapenet}, Section~\ref{sec:multiview_ebay}). Finally, we compare our approach with a Multi View Stereo method on reconstructing objects with various texture levels and viewpoint sparsity (Section~\ref{sec:mvs}).

\subsection{Dataset}
\label{sec:dataset}
\noindent\textbf{ShapeNet:} The ShapeNet dataset is a collection of 3D CAD models that are organized according to the WordNet hierarchy. We used a subset of the ShapeNet dataset which consists of 50,000 models and 13 major categories (see Table~\ref{fig:multiview_plot}(c) for a complete list). We split the dataset into training and testing sets, with 4/5 for training and the remaining 1/5 for testing. We refer to these two datasets as the ShapeNet training set and testing set throughout the experiments section. 

\noindent\textbf{PASCAL 3D:} The PASCAL 3D dataset is composed of PASCAL 2012 detection images augmented with 3D CAD model alignment~\cite{xiang2014beyond}.

\noindent\textbf{Online Products:} The dataset ~\cite{ebay} contains
images of 23,000 items sold online. MVS and SFM methods fail on these images due to ultra-wide baselines. Since the dataset does not have the ground-truth 3D CAD models, we only used the dataset for qualitative evaluation.

\noindent\textbf{MVS CAD Models:} To compare our method with a Multi View Stereo method \cite{cgstudio}, we collected 4 different categories of high-quality CAD models. All CAD models have texture-rich surfaces and were placed on top of a texture-rich paper to aid the camera localization of the MVS method.
\vspace{0.1cm}

\noindent\textit{Metrics:} We used two metrics in evaluating the reconstruction quality. The primary metric was the voxel Intersection-over-Union (IoU) between a 3D voxel reconstruction and its ground truth voxelized model. More formally,

\vspace{\alignvspace}
\begin{align}
IoU &= \sum_{i,j,k}\left[ I(p_{(i, j, k)} > t)I(y_{(i, j, k)}) \right]/\sum_{i,j,k}\left[ I\left(I(p_{(i, j, k)} > t) + I(y_{(i, j, k)})\right) \right]
\end{align}
\vspace{\alignvspace}

where variables are defined in Section~\ref{sec:loss}. $I(\cdot)$ is an indicator function and $t$ is a voxelization threshold. Higher IoU values indicate better reconstructions. We also report the cross-entropy loss (Section~\ref{sec:loss}) as a secondary
metric. Lower loss values indicate higher confidence reconstructions.

\subsection{Network Structures Comparison}
\label{sec:network_comparison}

\begin{table}[!b]
\vspace{-0.08cm}
\scriptsize
  \caption{Reconstruction performance of 3D-LSTM variations according to cross-entropy loss and IoU using 5 views.}
\begin{center}
  \begin{tabular}{ | c || c | c | c | c | c |}
    \hline
                  & Encoder  & Recurrence & Decoder  & Loss & IoU \\ \hline
    3D-LSTM-1     & simple   & LSTM       & simple   & 0.116         & 0.499 \\ \hline
    3D-GRU-1      & simple   & GRU        & simple   & 0.105         & 0.540 \\ \hline
    3D-LSTM-3     & simple   & LSTM       & simple   & 0.106         & 0.539 \\ \hline
    3D-GRU-3      & simple   & GRU        & simple   & 0.091         & 0.592 \\ \hline
    Res3D-GRU-3  & residual & GRU        & residual & \textbf{0.080}& \textbf{0.634} \\ \hline
  \end{tabular}
\end{center}
\label{table:network_comparison}
\end{table}

We tested 5 variations of our 3D-R2N2 as described in Section~\ref{sec:network}. The first four networks are based on the standard feed-forward CNN (top Fig.~\ref{fig:network}) and the fifth network is the residual network (bottom Fig.~\ref{fig:network}). For the first four networks, we used either GRU or LSTM units and and varied the convolution kernel to be either $1 \times 1 \times 1$ [3D-LSTM/GRU-3] or $3 \times 3 \times 3$ [3D-LSTM/GRU-3]. The residual network used GRU units and $3\times3\times3$ convolutions [Res3D-GRU-3]. These networks were trained on the ShapeNet training set and tested on the ShapeNet testing set. We used 5 views in the experiment. Table~\ref{table:network_comparison} shows the results. We observe that 1) the GRU-based networks outperform the LSTM-based networks, 2) that the networks with neighboring recurrent unit connections ($3\times3\times3$ convolutions) outperform the networks that have no neighboring recurrent unit connection ($1\times1\times1$ convolutions), and 3) that the deep residual network variation further boosts the reconstruction performance.

\vspace{\tablevspace}

\subsection{Single Real-World Image Reconstruction}
\label{sec:pascal}
We evaluated the performance of our network in single-view reconstruction using real-world images, comparing the performance with that of a recent method by Kar et al.~\cite{kar2015category}.
To make a quantitative comparison, we used images from the PASCAL VOC 2012
dataset~\cite{everingham2011pascal} and its corresponding 3D models from the PASCAL
3D+ dataset~\cite{xiang2014beyond}. We ran the experiments with the same
configuration as Kar et al. except that we allow the Kar et al. method to have ground-truth object segmentation
masks and keypoint labels as additional inputs for both training and testing.

\input{single_view_reconstruction_images}


\textbf{Training.} We fine-tuned a network trained on the ShapeNet dataset with
PASCAL 3D+. We used the PASCAL 3D+ validation set to find hyperparameters such as the number of fine-tuning iterations and
the voxelization threshold.

\textbf{Results.} As shown in Table~\ref{table:singlerealeval},
our approach outperforms the method of Kar et
al.~\cite{kar2015category} in every category. However, we observe that our network has some
difficulties reconstructing thin legs of chairs. Moreover, the network often
confuses thin flat panels with thick CRT screens when given a frontal view of the
monitor. Yet, our approach demonstrates a competitive quantitative
performance. For the qualitative results and comparisons, please see Fig.~\ref{fig:pascal}.

Aside from better performance, our network has several advantages over Kar et al.~\cite{kar2015category}. First, we do not need to train and test 
per-category. Our network trains and reconstructs without knowing the
object category. Second, our network does not require object
segmentation masks and keypoint labels as additional inputs. Kar et al. does
demonstrate the possibility of testing on a wild unlabeled image by estimating
the segmentation and keypoints. However, our network outperforms their method
tested with ground truth labels.

\begin{table}[htp!]
\vspace{\tablevspace}
\scriptsize
  \caption{Per-category reconstruction of PASCAL VOC compared using voxel
    Intersection-over-Union (IoU). Note that the experiments were ran with the same
    configuration except that the method of Kar et al.~\cite{kar2015category} took
  ground-truth object segmentation masks and keypoint labels as additional
inputs for both training and testing.}
\begin{center}
  \begin{tabular}{ | c || c | c | c | c | c | c | c | c | c | c || c | }
    \hline
    & \textbf{aero} & \textbf{bike} & \textbf{boat} & \textbf{bus} & \textbf{car} & \textbf{chair} & \textbf{mbike} & \textbf{sofa} & \textbf{train} & \textbf{tv} & \textbf{mean} \\ \hline
    Kar et al.~\cite{kar2015category} & 0.298 & 0.144 & 0.188 & 0.501 & 0.472 & 0.234 & 0.361 & 0.149 & 0.249 & 0.492 & 0.318 \\ \hline
    ours [LSTM-1]& 0.472 & 0.330 & 0.466 & 0.677 & 0.579 & 0.203 & 0.474 & 0.251 & 0.518 & 0.438 & 0.456 \\ \hline
    ours [Res3D-GRU-3]& \textbf{0.544} & \textbf{0.499} & \textbf{0.560} & \textbf{0.816} & \textbf{0.699} & \textbf{0.280} & \textbf{0.649} & \textbf{0.332} & \textbf{0.672} & \textbf{0.574} & \textbf{0.571} \\ \hline
  \end{tabular}
\end{center}
\label{table:singlerealeval}
\vspace{\tablevspace}
\vspace{\tablevspace}
\vspace{\tablevspace}
\end{table}

\subsection{Multi-view Reconstruction Evaluation}
\label{sec:multiview_shapenet}
In this section, we report a quantitative evaluation of our network's performance in multi-view reconstruction on the ShapeNet testing set.

\textbf{Experiment setup.}
We used the [Res3D-GRU-3] network in this experiment. We evaluated the network with the ShapeNet testing set. The testing set consisted of 8725 models in 13 major categories. We rendered five random views for each
model, and we applied a uniform colored background to the image. 
We report both softmax loss and
intersection over union(IoU) with a voxelization threshold of 0.4 between the predicted and the ground truth voxel models. 

\begin{figure}[!t]
\begin{center}
    \begin{tabular}{cc}
        \begin{minipage}{.6\linewidth}
            \begin{tabular}{cc}
              \includegraphics[width=0.5\linewidth]{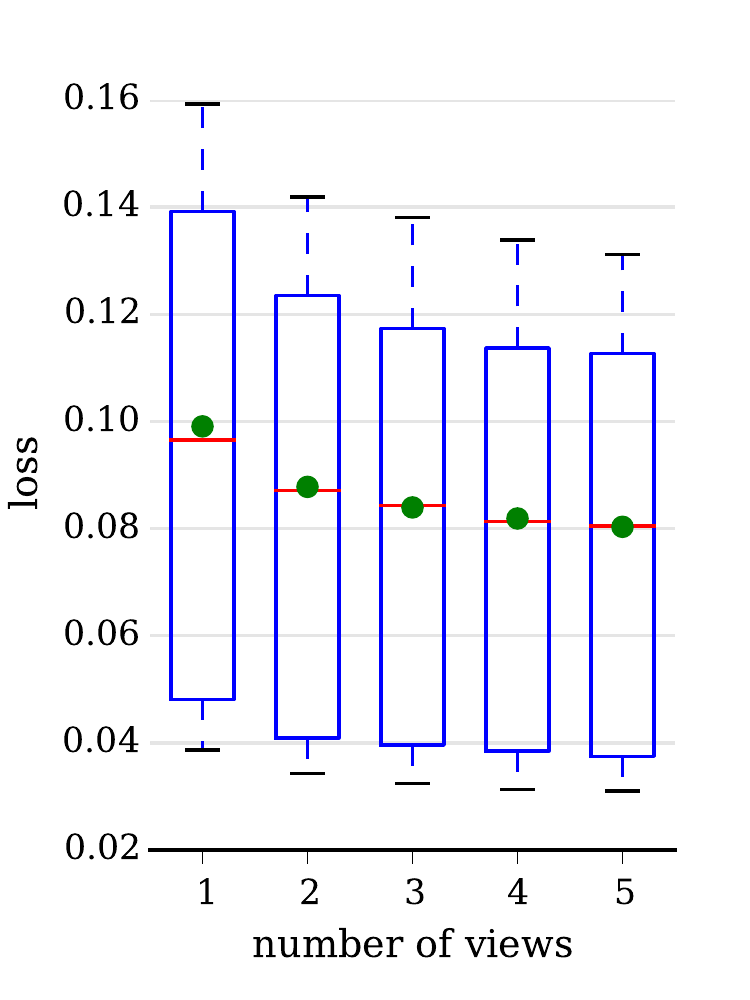} &
              \hspace{-0.5cm}
              \includegraphics[width=0.5\linewidth]{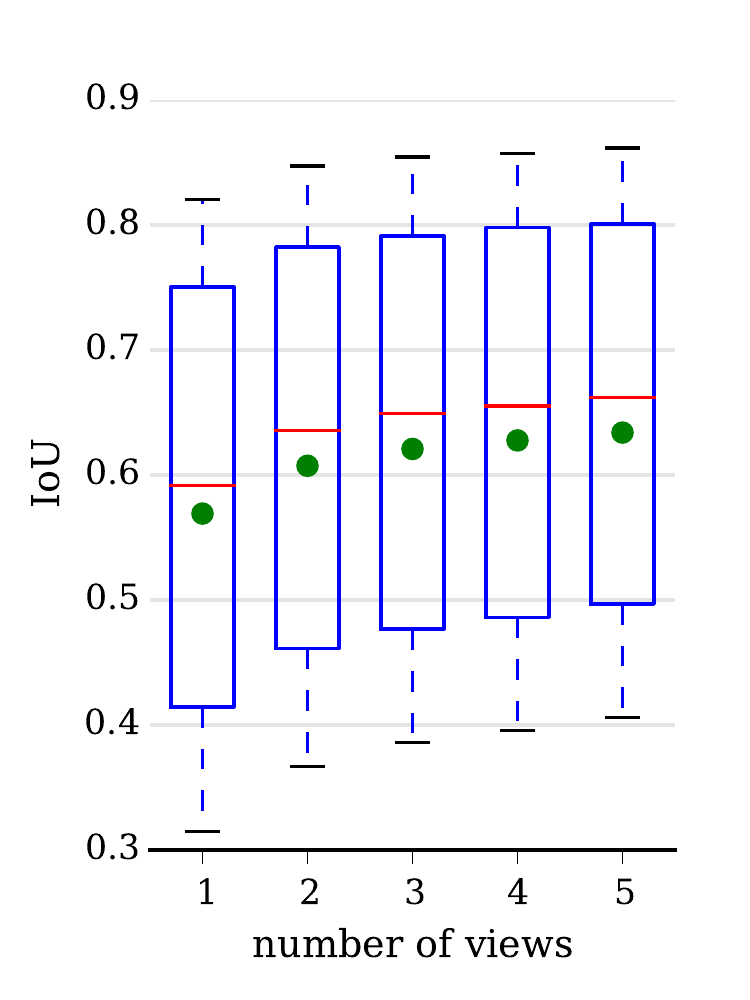} \\
              (a) Cross entropy loss & (b) Voxel IoU
            \end{tabular}
        \end{minipage} & 
        
        \hspace{-1cm}
        \begin{minipage}{.5\linewidth}
            \vspace{0.3cm}
            \begin{center}
            \begin{center}
                \scriptsize
                \begin{tabular}{ c | c | c | c | c | c}
                \hline
                \# views & 1 & 2 & 3 & 4 & 5 \\ \hline
                plane & 0.513 & 0.536 & 0.549 & 0.556 & 0.561 \\
                bench & 0.421 & 0.484 & 0.502 & 0.516 & 0.527 \\
                cabinet & 0.716 & 0.746 & 0.763 & 0.767 & 0.772 \\
                car & 0.798 & 0.821 & 0.829 & 0.833 & 0.836 \\
                chair & 0.466 & 0.515 & 0.533 & 0.541 & 0.550 \\
                monitor & 0.468 & 0.527 & 0.545 & 0.558 & 0.565 \\
                lamp & 0.381 & 0.406 & 0.415 & 0.416 & 0.421 \\
                speaker & 0.662 & 0.696 & 0.708 & 0.714 & 0.717 \\
                firearm & 0.544 & 0.582 & 0.593 & 0.595 & 0.600 \\
                couch & 0.628 & 0.677 & 0.690 & 0.698 & 0.706 \\
                table & 0.513 & 0.550 & 0.564 & 0.573 & 0.580 \\
                cellphone & 0.661 & 0.717 & 0.732 & 0.738 & 0.754 \\
                watercraft & 0.513 & 0.576 & 0.596 & 0.604 & 0.610 \\
                \hline
                \end{tabular}\\[0.6cm]
            \end{center}
            (c) Per-category IoU
            \end{center}
        \end{minipage}
    \end{tabular}
\end{center}
\vspace{\figurevspace}
\caption{(a), (b): Multi-view reconstruction using our model on the ShapeNet dataset. The performance is reported in median (red line) and mean (green dot) cross-entropy loss and intersection over union (IoU) values. The box plot shows 25\% and 75\%, with caps showing 15\% and 85\%. (c): Per-category reconstruction of the ShapeNet dataset using our model. The values are average IoU.}
\label{fig:multiview_plot}
\vspace{\figurevspace}
\end{figure}

\textbf{Overall results.} We first investigate the quality of the reconstructed
models under different numbers of views.  Fig.
\ref{fig:multiview_plot}(a) and (b) show that
reconstruction quality improves as the number of views increases. The fact that the
marginal gain decreases accords with our assumption that each
additional view provides less information since two random views are very likely
to have partial overlap.

\textbf{Per-category results.} We also report the reconstruction IoUs on each of the 13 major categories in the testing set in Table~\ref{fig:multiview_plot}. We observed that the reconstruction quality improved for 
every category as the number of views increased, but the quality varied depending on the category. 
Cabinets, cars, and speakers had the highest
reconstruction performance since the objects are bulky-shaped and
have less (shape) variance compared to other classes. The network performed worse on the lamp, bench, and table categories. These classes have much higher shape
variation than the other classes. For example, a lamp can have a slim arm or a large lampshade 
which may move around, and chairs and tables have various types of supporting structures.

\textbf{Qualitative results.} Fig.~\ref{fig:multiview_all}(a) shows some sample reconstructions from the ShapeNet testing set.
One exemplary instance is the truck shown in row 2. In the initial view, only the front part of the truck is visible. 
The network took the safest guess that the object is a sedan, which is the most common shape in the car category. 
Then the network produced a more accurate reconstruction of the truck after seeing more views. 
All other instances show similar improvements as the network sees more views of the objects. Fig.~\ref{fig:multiview_all}(b) shows
 two failure cases.

\subsection{Reconstructing Real World Images}
\label{sec:multiview_ebay}
In this experiment, we tested our network on the Online Products dataset for qualitative evaluation. Images that were not square-shaped were padded with white pixels. 

Fig.~\ref{fig:multiview_all}(c) shows some sample reconstructions. The result shows that the network is capable of reconstructing
real world objects using only synthetic data as training samples. It also demonstrates that the network improves the
reconstructions after seeing additional views of the objects.  One exemplary instance is the reconstruction of couch as shown in row 1. 
The initial side view of the couch led the network to believe that it was a one-seater sofa, but after seeing the front of the couch, the 
network immediately refined its reconstruction to reflect the observation. Similar behaviors are also shown in other samples. Some failure cases are shown in Fig.\ref{fig:multiview_all}(d).

\input{figures/multiview_all}

\subsection{Multi View Stereo(MVS) vs. 3D-R2N2}
\label{sec:mvs}

In this experiment, we compare our approach with a MVS method on reconstructing objects that are of various texture levels with different number of views. MVS methods are limited by the accuracy of feature correspondences across different views. Therefore, they tend to fail reconstructing textureless objects or images from sparsely positioned camera viewpoints. In contrast, our method does not require accurate feature correspondences or adjacent camera viewpoints.

\begin{figure}
\vspace{\tablevspace}
\centering
\scriptsize
    \begin{tabular}{ccccc}
      \includegraphics[width=0.15\linewidth]{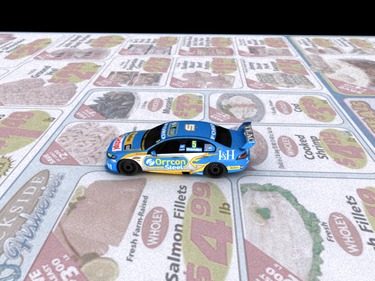} &
      \includegraphics[width=0.15\linewidth]{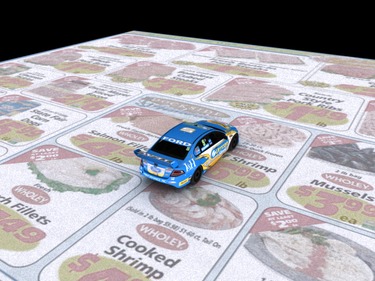} &
      \includegraphics[width=0.15\linewidth]{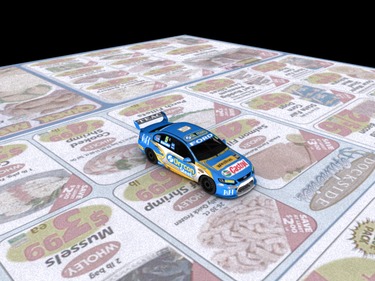} &
      \includegraphics[width=0.15\linewidth]{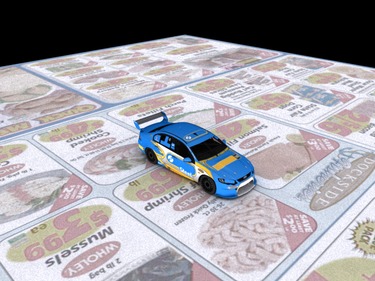} &
      \includegraphics[width=0.15\linewidth]{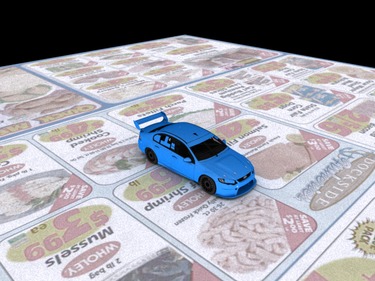} \\
      (a) & (b) & (c) & (d) & (e)
    \end{tabular}
\caption{Rendered images with various viewpoints (a,b,c) and texture levels (c, d, e) (from high to low), used for the comparison experiment against MVS~\cite{mvs}.}
\vspace{\tablevspace}
\label{fig:mvs_inputs}
\end{figure}

\noindent\textbf{Dataset:} We used high-quality CAD models of 4 different categories and augmented their texture strengths to low, medium and high by manually editing their textures. We then rendered the models from viewpoints with uniformly sampled azimuth angles. Please refer to Fig.~\ref{fig:mvs_inputs} for some samples of rendered models across different viewpoints and texture strengths. For each texture level and number of views configuration, both the MVS method and our network took identical sets of images as inputs.

\noindent\textbf{Experiment setup.} We used a Patch-Match~\cite{barnes2009patchmatch}-based off-the-shelf implementation~\cite{mvs} as the MVS method. The MVS method takes images along with their camera positions estimated by Global SFM~\cite{moulon2013global} and outputs the reconstructed model. For our network, we used the [Res3D-GRU-3] network trained with at most 5 views. In order to cope with more views, we fine-tuned our network with samples that have a maximum of 24 views for 5000 iterations using the ShapeNet training set. We quantified the quality of the reconstructions using IoU of the voxels. The network was voxelized with the occupancy probability threshold set to 0.1. The mesh reconstructed from the MVS method was voxelized into a $32\times32\times32$ grid for comparison.

\begin{figure}
\vspace{\tablevspace}
\vspace{\tablevspace}
\centering
\scriptsize
\begin{tabular}{MMMMM}
\multirow{3}{*}{\includegraphics[width=0.25\linewidth]{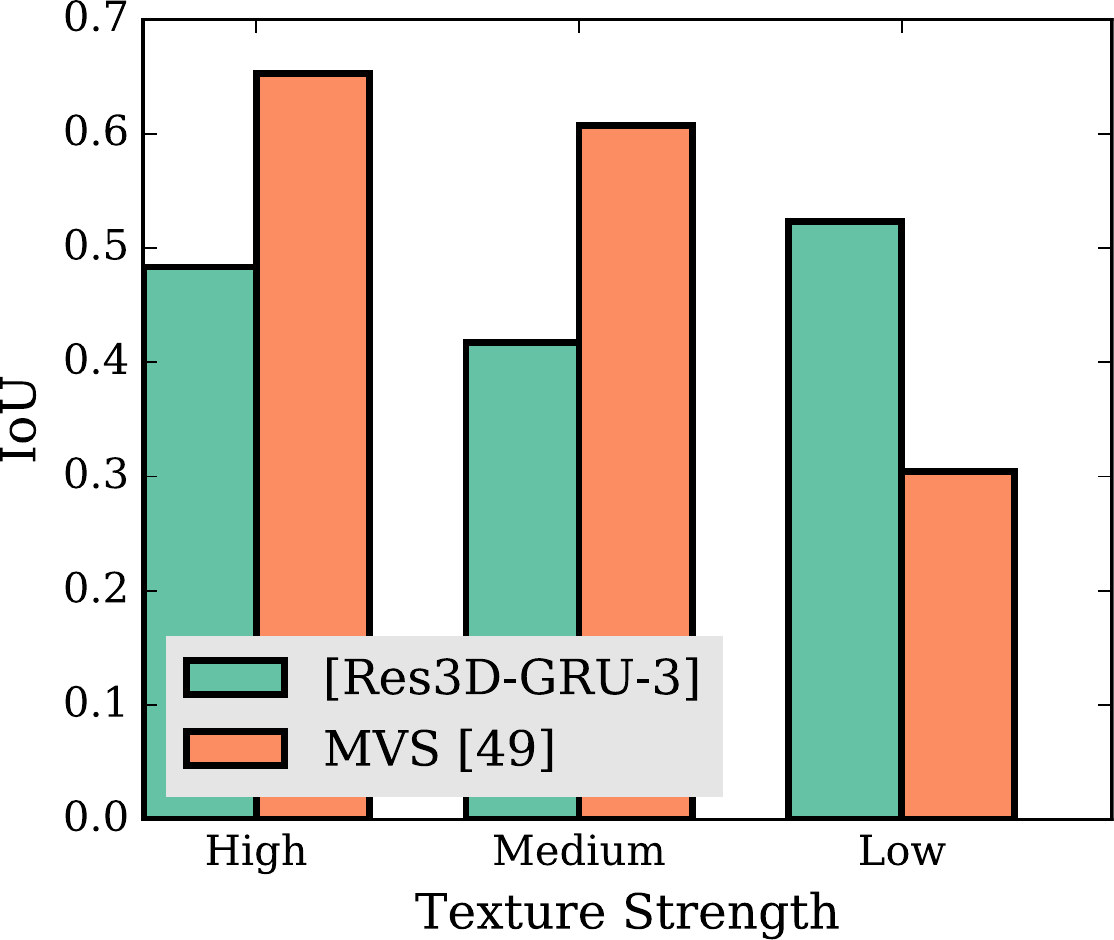}} &
\multirow{3}{*}{\includegraphics[width=0.25\linewidth]{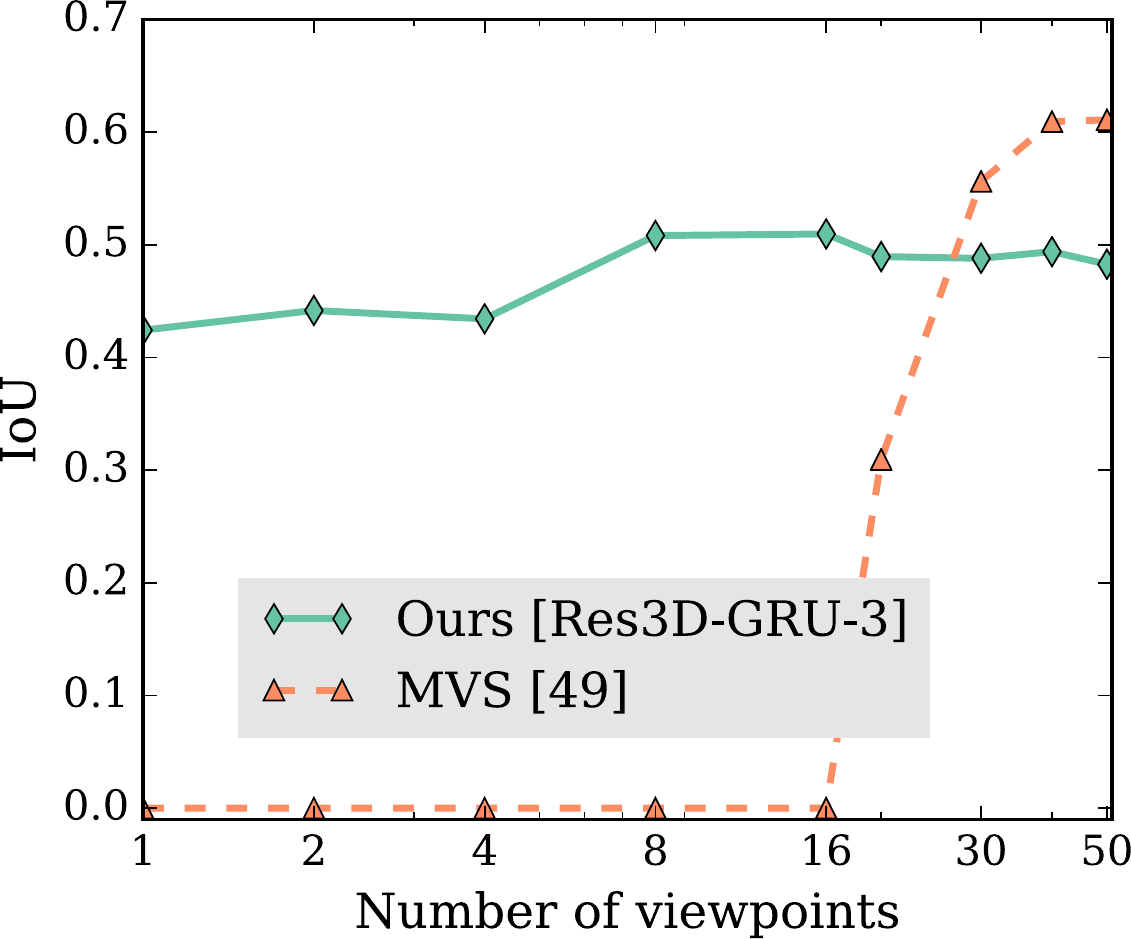}} &
\includegraphics[width=0.15\linewidth]{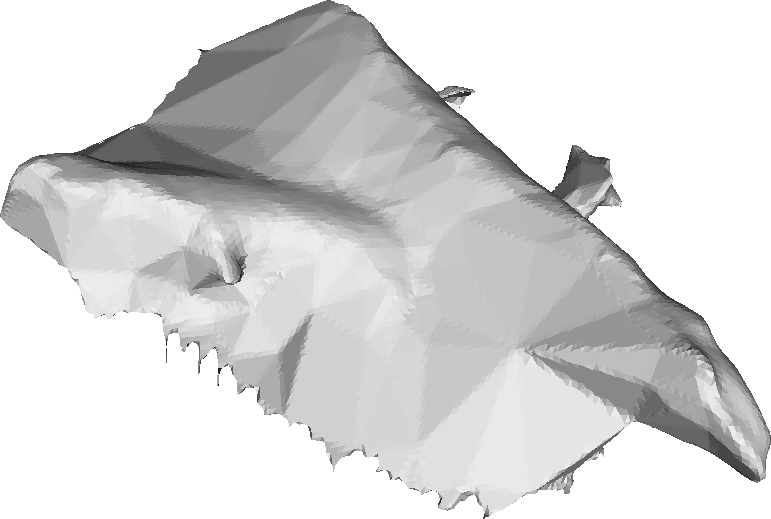} &
\includegraphics[width=0.15\linewidth]{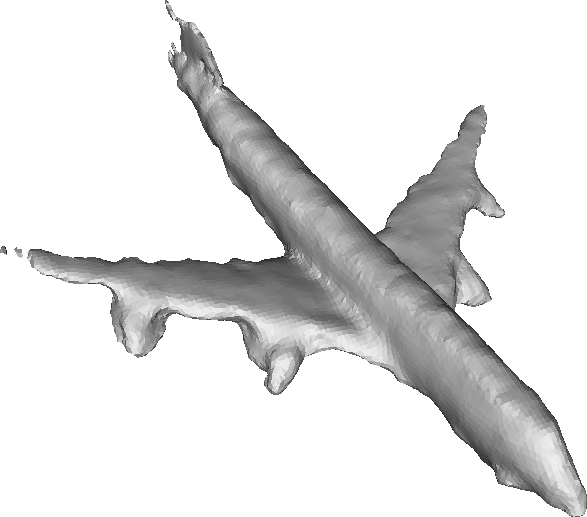} &
\includegraphics[width=0.15\linewidth]{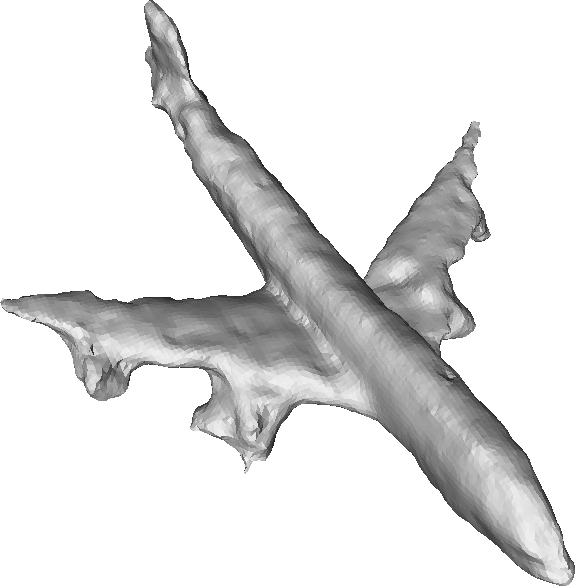}\tabularnewline
 &  & (c) & (d) & (e)\tabularnewline
 &  & 
 \includegraphics[width=0.15\linewidth]{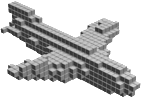} &
 \includegraphics[width=0.15\linewidth]{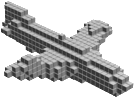} &
 \includegraphics[width=0.15\linewidth]{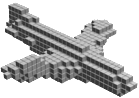}\tabularnewline
(a) & (b) & (f) & (g) & (h)
\tabularnewline
\end{tabular}
\caption{Reconstruction performance of MVS~\cite{mvs} compared with that of our network. (a) shows how texture strengths affect the reconstructions of MVS and our network, averaged over 20, 30, 40, and 50 input views of all classes. (b) compares the quality of the reconstruction across the number of input images, averaged over all texture levels of all classes. (c-e) show the reconstruction result of MVS and (f-h) show the reconstruction results from our method [Res3D-GRU-3] on a high-texture airplane model with 20, 30, and 40 input views respectively.}
\label{fig:sfm}
\vspace{\tablevspace}
\vspace{\tablevspace}
\end{figure}

\noindent\textbf{Results.} The results are shown in Fig.~\ref{fig:sfm} (a) and (b).  We observed 1) that our model worked with as few as one view, whereas the MVS method failed completely when the number of views was less than 20 (IoU=0), and 2) that our model worked regardless of the objects' texture level, whereas the MVS method frequently failed to reconstruct objects that had low texture level even when a large number of views were provided.
This shows that our method works in situations where MVS methods would perform poorly or completely fail. Note that the reconstruction performance of our method decreased after the number of views passed 24. This is because we only fine-tuned our network on samples with a maximum of 24 views.

We also discovered some limitations of our method. First, our method could not reconstruct as many details as the MVS method did when given more than 30 different views of the model. Second, our method performed worse in reconstructing objects with high texture levels. This is largely because most models in the ShapeNet training set have low texture level.

%% file: single_view_reconstruction_images.tex
\newcommand\wx{0.09}
\begin{figure}[htp!]
    \vspace{\figurevspace}
    \centering
    \begin{tabular}{ccccccccc}
        &\tiny{Input} & \tiny{Ground Truth} & \tiny{Ours} & \tiny{Kar et al.~\cite{kar2015category}} & \tiny{Input} & \tiny{Ground Truth} & \tiny{Ours} & \tiny{Kar et al.~\cite{kar2015category}}\\
      &\includegraphics[height=\wx\linewidth,width=\wx\linewidth,keepaspectratio]{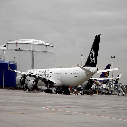} &  
      \includegraphics[height=\wx\linewidth,width=\wx\linewidth,keepaspectratio]{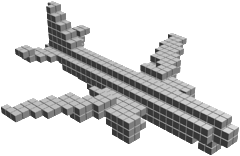} & 
      \includegraphics[height=\wx\linewidth,width=\wx\linewidth,keepaspectratio]{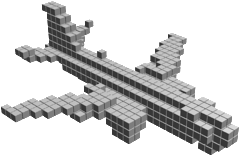} &   
      \includegraphics[height=\wx\linewidth,width=\wx\linewidth,keepaspectratio]{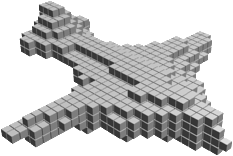} & 
      \includegraphics[height=\wx\linewidth,width=\wx\linewidth,keepaspectratio]{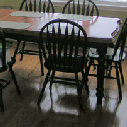} &
      \includegraphics[height=\wx\linewidth,width=\wx\linewidth,keepaspectratio]{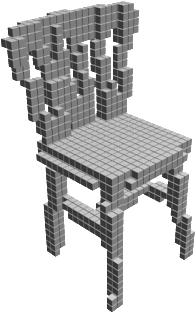} &   
      \includegraphics[height=\wx\linewidth,width=\wx\linewidth,keepaspectratio]{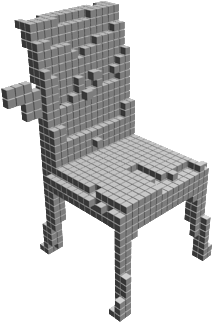} & 
      \includegraphics[height=\wx\linewidth,width=\wx\linewidth,keepaspectratio]{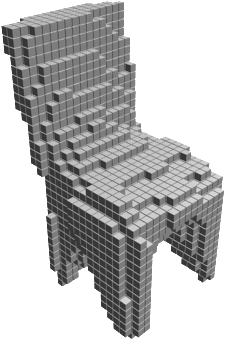} \\
      (a) &\includegraphics[height=\wx\linewidth,width=\wx\linewidth,keepaspectratio]{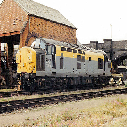} &  
      \includegraphics[height=\wx\linewidth,width=\wx\linewidth,keepaspectratio]{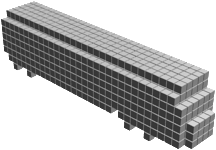} & 
      \includegraphics[height=\wx\linewidth,width=\wx\linewidth,keepaspectratio]{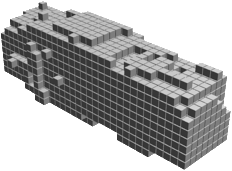} &   
      \includegraphics[height=\wx\linewidth,width=\wx\linewidth,keepaspectratio]{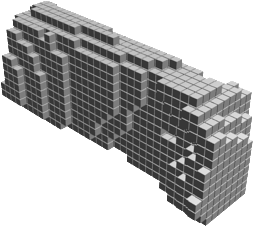} & 
      \includegraphics[height=\wx\linewidth,width=\wx\linewidth,keepaspectratio]{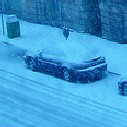} &
      \includegraphics[height=\wx\linewidth,width=\wx\linewidth,keepaspectratio]{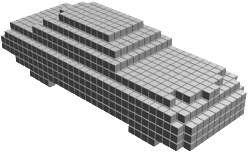} &   
      \includegraphics[height=\wx\linewidth,width=\wx\linewidth,keepaspectratio]{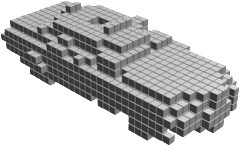} & 
      \includegraphics[height=\wx\linewidth,width=\wx\linewidth,keepaspectratio]{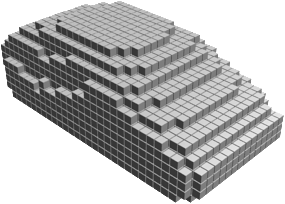} \\
      &\includegraphics[height=\wx\linewidth,width=\wx\linewidth,keepaspectratio]{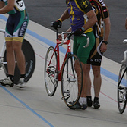} &  
      \includegraphics[height=\wx\linewidth,width=\wx\linewidth,keepaspectratio]{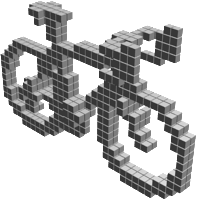} & 
      \includegraphics[height=\wx\linewidth,width=\wx\linewidth,keepaspectratio]{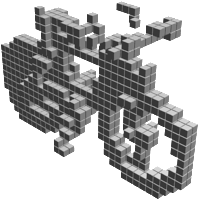} &   
      \includegraphics[height=\wx\linewidth,width=\wx\linewidth,keepaspectratio]{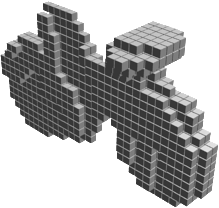} & 
      \includegraphics[height=\wx\linewidth,width=\wx\linewidth,keepaspectratio]{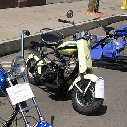} &
      \includegraphics[height=\wx\linewidth,width=\wx\linewidth,keepaspectratio]{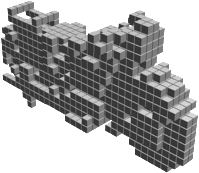} &   
      \includegraphics[height=\wx\linewidth,width=\wx\linewidth,keepaspectratio]{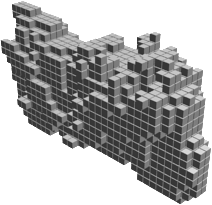} & 
      \includegraphics[height=\wx\linewidth,width=\wx\linewidth,keepaspectratio]{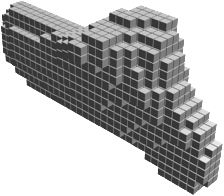} \\
      &\includegraphics[height=\wx\linewidth,width=\wx\linewidth,keepaspectratio]{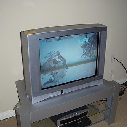} &  
      \includegraphics[height=\wx\linewidth,width=\wx\linewidth,keepaspectratio]{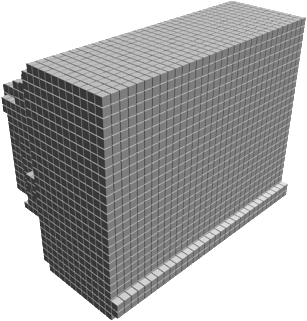} & 
      \includegraphics[height=\wx\linewidth,width=\wx\linewidth,keepaspectratio]{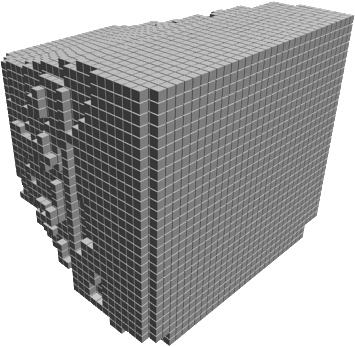} &   
      \includegraphics[height=\wx\linewidth,width=\wx\linewidth,keepaspectratio]{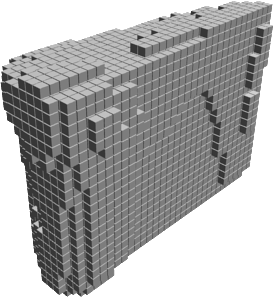} & 
      \includegraphics[height=\wx\linewidth,width=\wx\linewidth,keepaspectratio]{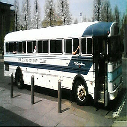} &
      \includegraphics[height=\wx\linewidth,width=\wx\linewidth,keepaspectratio]{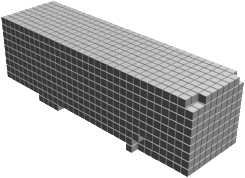} &   
      \includegraphics[height=\wx\linewidth,width=\wx\linewidth,keepaspectratio]{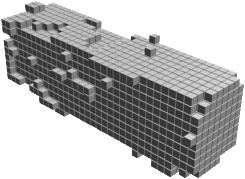} & 
      \includegraphics[height=\wx\linewidth,width=\wx\linewidth,keepaspectratio]{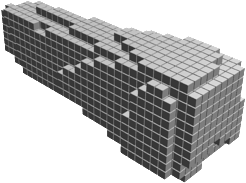} \\

      \hline
    \end{tabular}
    
    \begin{tabular}{ccccccccc}
       & \tiny{Input} & \tiny{Ground Truth} & \tiny{Ours} & \tiny{Kar et al.~\cite{kar2015category}} & \tiny{Input} & \tiny{Ground Truth} & \tiny{Ours} & \tiny{Kar et al.~\cite{kar2015category}}\\
      (b) &\includegraphics[height=\wx\linewidth,width=\wx\linewidth,keepaspectratio]{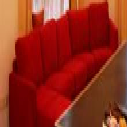} &  
      \includegraphics[height=\wx\linewidth,width=\wx\linewidth,keepaspectratio]{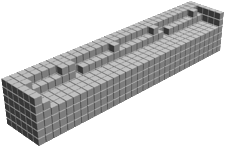} & 
      \includegraphics[height=\wx\linewidth,width=\wx\linewidth,keepaspectratio]{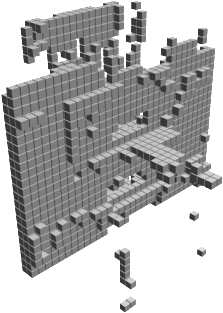} &   
      \includegraphics[height=\wx\linewidth,width=\wx\linewidth,keepaspectratio]{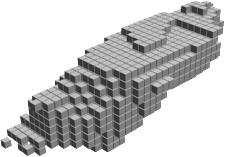} & 
      \includegraphics[height=\wx\linewidth,width=\wx\linewidth,keepaspectratio]{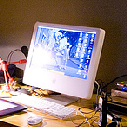} &
      \includegraphics[height=\wx\linewidth,width=\wx\linewidth,keepaspectratio]{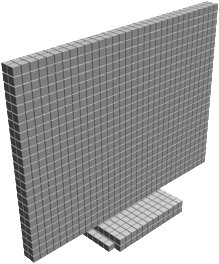} &   
      \includegraphics[height=\wx\linewidth,width=\wx\linewidth,keepaspectratio]{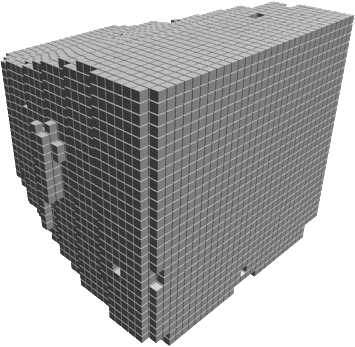} & 
      \includegraphics[height=\wx\linewidth,width=\wx\linewidth,keepaspectratio]{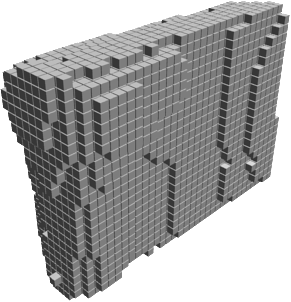} \\
      &\includegraphics[height=\wx\linewidth,width=\wx\linewidth,keepaspectratio]{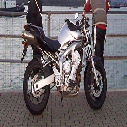} &  
      \includegraphics[height=\wx\linewidth,width=\wx\linewidth,keepaspectratio]{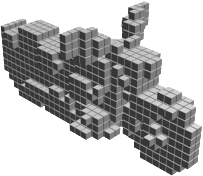} & 
      \includegraphics[height=\wx\linewidth,width=\wx\linewidth,keepaspectratio]{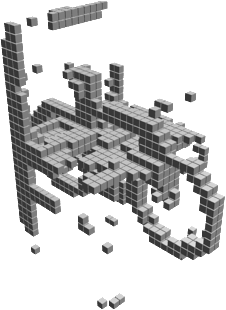} &   
      \includegraphics[height=\wx\linewidth,width=\wx\linewidth,keepaspectratio]{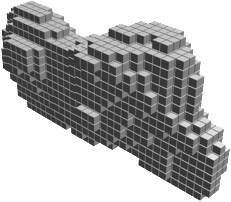} & 
      \includegraphics[height=\wx\linewidth,width=\wx\linewidth,keepaspectratio]{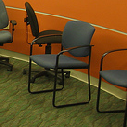} &
      \includegraphics[height=\wx\linewidth,width=\wx\linewidth,keepaspectratio]{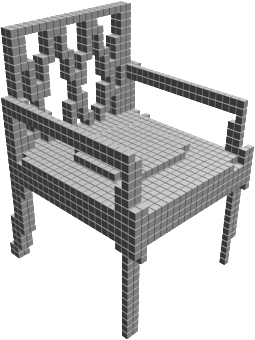} &   
      \includegraphics[height=\wx\linewidth,width=\wx\linewidth,keepaspectratio]{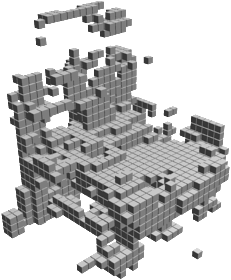} & 
      \includegraphics[height=\wx\linewidth,width=\wx\linewidth,keepaspectratio]{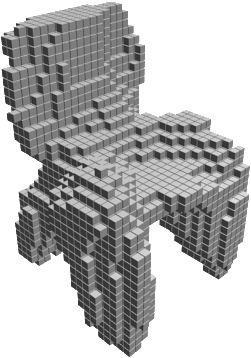} \\

    \end{tabular}
    \caption{(a) Reconstruction samples of PASCAL VOC dataset. (b) Failed reconstructions on the PASCAL VOC dataset. Note that Kar et al.~\cite{kar2015category} is trained/tested per category and takes ground-truth object segmentation masks and keypoint labels as additional input.}
    \label{fig:pascal}
    \vspace{\figurevspace}
\end{figure}

%% file: figures/multiview_all.tex
\newcommand\wxsn{0.075}
\newcommand\wxsnfail{0.07}
\begin{figure}
\begin{center}
\begin{tabular}{ccccccccccccc}
  & \includegraphics[width=\wxsn\linewidth,height=\wxsn\linewidth,keepaspectratio]{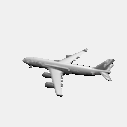} &  
  \includegraphics[width=\wxsn\linewidth,height=\wxsn\linewidth,keepaspectratio]{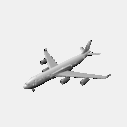} &  
  \includegraphics[width=\wxsn\linewidth,height=\wxsn\linewidth,keepaspectratio]{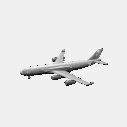} &  
  \includegraphics[width=\wxsn\linewidth,height=\wxsn\linewidth,keepaspectratio]{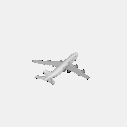} &  
  \includegraphics[width=\wxsn\linewidth,height=\wxsn\linewidth,keepaspectratio]{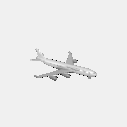} &
  &
  \includegraphics[width=\wxsn\linewidth,height=\wxsn\linewidth,keepaspectratio]{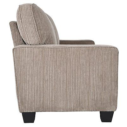} &  
  \includegraphics[width=\wxsn\linewidth,height=\wxsn\linewidth,keepaspectratio]{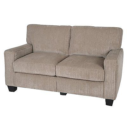} &
  \includegraphics[width=\wxsn\linewidth,height=\wxsn\linewidth,keepaspectratio]{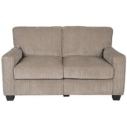} &  
  \includegraphics[width=\wxsn\linewidth,height=\wxsn\linewidth,keepaspectratio]{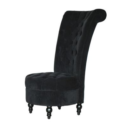} &  
  \includegraphics[width=\wxsn\linewidth,height=\wxsn\linewidth,keepaspectratio]{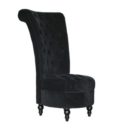} &  
  \includegraphics[width=\wxsn\linewidth,height=\wxsn\linewidth,keepaspectratio]{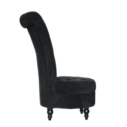} \\
  
  &\includegraphics[width=\wxsn\linewidth,height=\wxsn\linewidth,keepaspectratio]{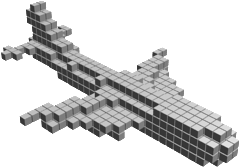} &  
  \includegraphics[width=\wxsn\linewidth,height=\wxsn\linewidth,keepaspectratio]{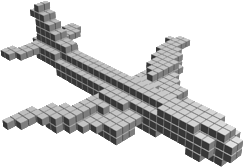} &  
  \includegraphics[width=\wxsn\linewidth,height=\wxsn\linewidth,keepaspectratio]{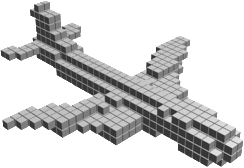} &  
  \includegraphics[width=\wxsn\linewidth,height=\wxsn\linewidth,keepaspectratio]{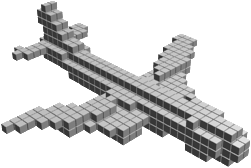} &  
  \includegraphics[width=\wxsn\linewidth,height=\wxsn\linewidth,keepaspectratio]{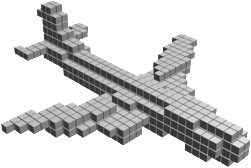} &
  &
  \includegraphics[width=\wxsn\linewidth,height=\wxsn\linewidth,keepaspectratio]{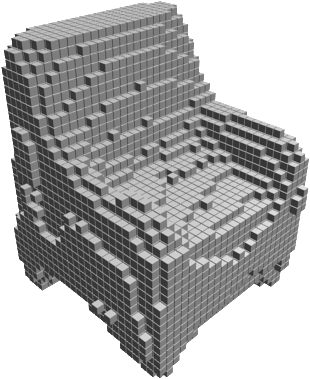} &  
  \includegraphics[width=\wxsn\linewidth,height=\wxsn\linewidth,keepaspectratio]{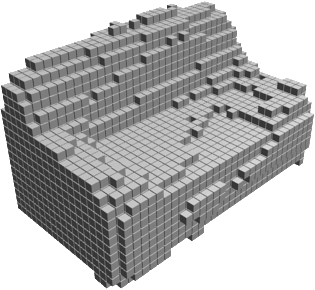} &  
  \includegraphics[width=\wxsn\linewidth,height=\wxsn\linewidth,keepaspectratio]{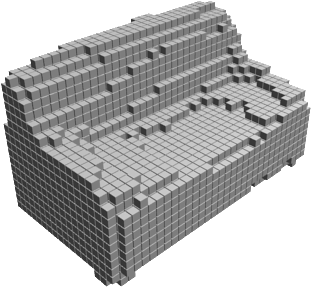} &  
  \includegraphics[width=\wxsn\linewidth,height=\wxsn\linewidth,keepaspectratio]{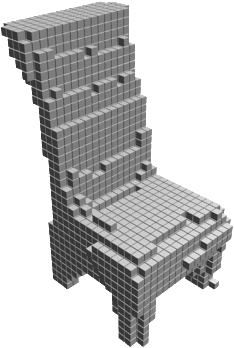} &  
  \includegraphics[width=\wxsn\linewidth,height=\wxsn\linewidth,keepaspectratio]{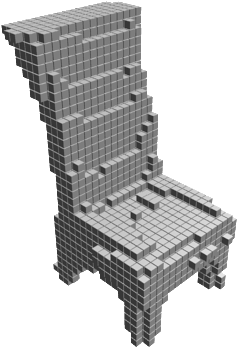} &  
  \includegraphics[width=\wxsn\linewidth,height=\wxsn\linewidth,keepaspectratio]{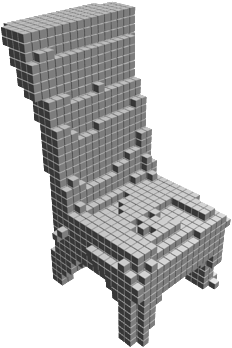} \\

  (a) &\includegraphics[width=\wxsn\linewidth,height=\wxsn\linewidth,keepaspectratio]{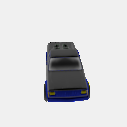} &  
  \includegraphics[width=\wxsn\linewidth,height=\wxsn\linewidth,keepaspectratio]{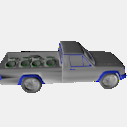} &  
  \includegraphics[width=\wxsn\linewidth,height=\wxsn\linewidth,keepaspectratio]{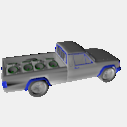} &  
  \includegraphics[width=\wxsn\linewidth,height=\wxsn\linewidth,keepaspectratio]{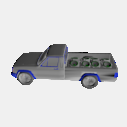} &  
  \includegraphics[width=\wxsn\linewidth,height=\wxsn\linewidth,keepaspectratio]{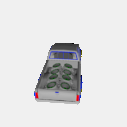} &
  (c) &
  \includegraphics[width=\wxsn\linewidth,height=\wxsn\linewidth,keepaspectratio]{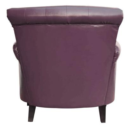} &  
  \includegraphics[width=\wxsn\linewidth,height=\wxsn\linewidth,keepaspectratio]{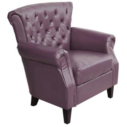} &  
  \includegraphics[width=\wxsn\linewidth,height=\wxsn\linewidth,keepaspectratio]{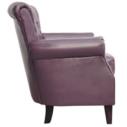} &  
  \includegraphics[width=\wxsn\linewidth,height=\wxsn\linewidth,keepaspectratio]{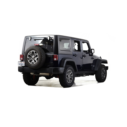} &  
  \includegraphics[width=\wxsn\linewidth,height=\wxsn\linewidth,keepaspectratio]{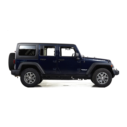} &  
  \includegraphics[width=\wxsn\linewidth,height=\wxsn\linewidth,keepaspectratio]{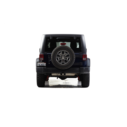} \\
  
  &\includegraphics[width=\wxsn\linewidth,height=\wxsn\linewidth,keepaspectratio]{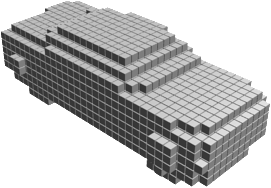} &  
  \includegraphics[width=\wxsn\linewidth,height=\wxsn\linewidth,keepaspectratio]{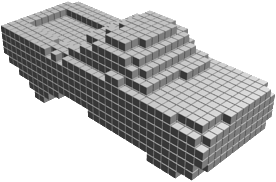} &  
  \includegraphics[width=\wxsn\linewidth,height=\wxsn\linewidth,keepaspectratio]{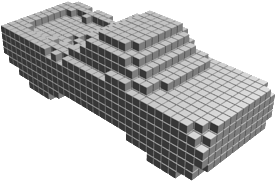} &  
  \includegraphics[width=\wxsn\linewidth,height=\wxsn\linewidth,keepaspectratio]{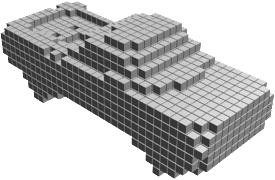} &  
  \includegraphics[width=\wxsn\linewidth,height=\wxsn\linewidth,keepaspectratio]{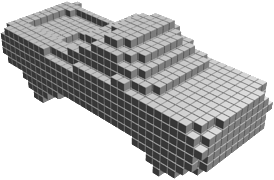} &
  &
  \includegraphics[width=\wxsn\linewidth,height=\wxsn\linewidth,keepaspectratio]{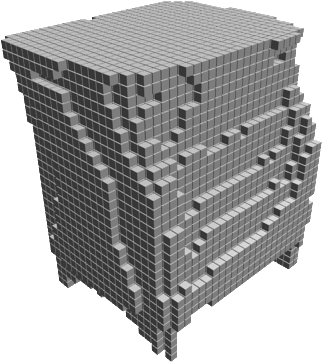} &  
  \includegraphics[width=\wxsn\linewidth,height=\wxsn\linewidth,keepaspectratio]{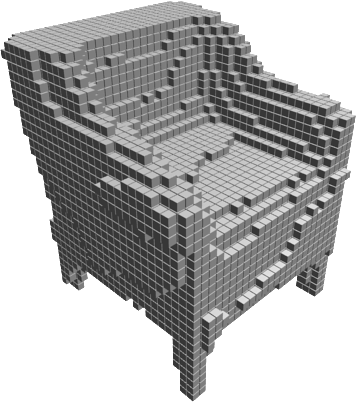} &  
  \includegraphics[width=\wxsn\linewidth,height=\wxsn\linewidth,keepaspectratio]{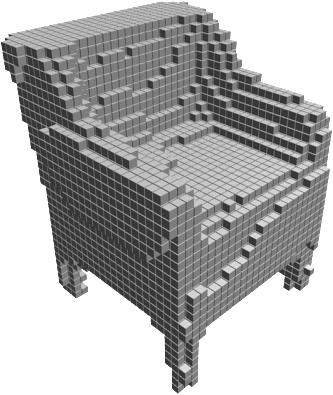} &  
  \includegraphics[width=\wxsn\linewidth,height=\wxsn\linewidth,keepaspectratio]{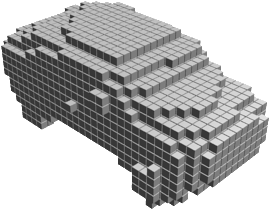} &  
  \includegraphics[width=\wxsn\linewidth,height=\wxsn\linewidth,keepaspectratio]{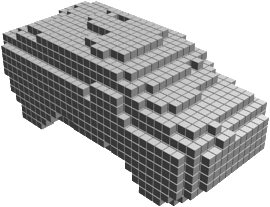} &  
  \includegraphics[width=\wxsn\linewidth,height=\wxsn\linewidth,keepaspectratio]{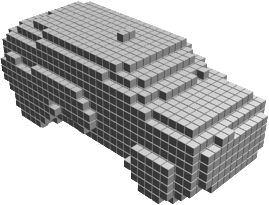} \\

  &\includegraphics[width=\wxsn\linewidth,height=\wxsn\linewidth,keepaspectratio]{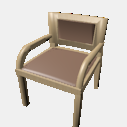} &  
  \includegraphics[width=\wxsn\linewidth,height=\wxsn\linewidth,keepaspectratio]{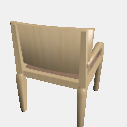} &  
  \includegraphics[width=\wxsn\linewidth,height=\wxsn\linewidth,keepaspectratio]{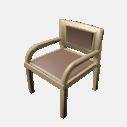} &  
  \includegraphics[width=\wxsn\linewidth,height=\wxsn\linewidth,keepaspectratio]{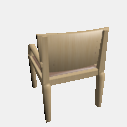} &  
  \includegraphics[width=\wxsn\linewidth,height=\wxsn\linewidth,keepaspectratio]{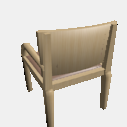} &
  &
  \includegraphics[width=\wxsn\linewidth,height=\wxsn\linewidth,keepaspectratio]{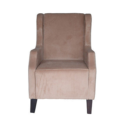} &  
  \includegraphics[width=\wxsn\linewidth,height=\wxsn\linewidth,keepaspectratio]{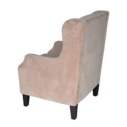} &  
  \includegraphics[width=\wxsn\linewidth,height=\wxsn\linewidth,keepaspectratio]{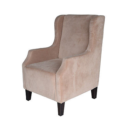} & 
  \includegraphics[width=\wxsn\linewidth,height=\wxsn\linewidth,keepaspectratio]{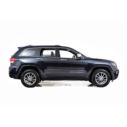} &  
  \includegraphics[width=\wxsn\linewidth,height=\wxsn\linewidth,keepaspectratio]{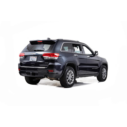} &  
  \includegraphics[width=\wxsn\linewidth,height=\wxsn\linewidth,keepaspectratio]{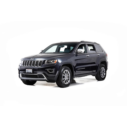} \\

  &\includegraphics[width=\wxsn\linewidth,height=\wxsn\linewidth,keepaspectratio]{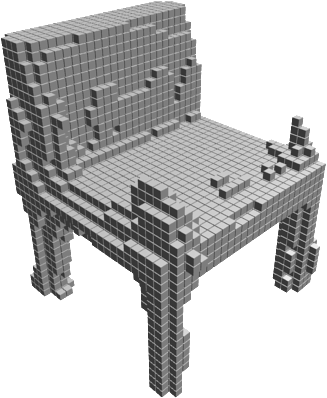} &  
  \includegraphics[width=\wxsn\linewidth,height=\wxsn\linewidth,keepaspectratio]{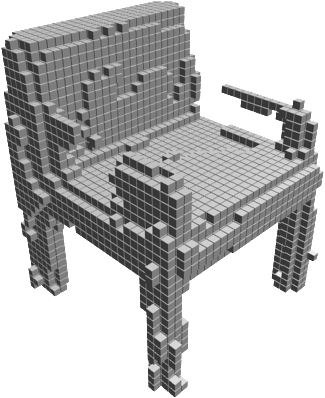} &  
  \includegraphics[width=\wxsn\linewidth,height=\wxsn\linewidth,keepaspectratio]{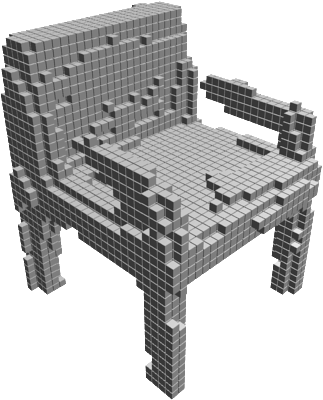} &  
  \includegraphics[width=\wxsn\linewidth,height=\wxsn\linewidth,keepaspectratio]{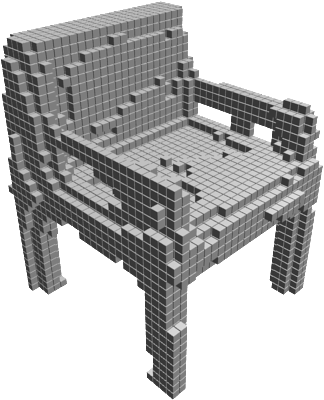} &  
  \includegraphics[width=\wxsn\linewidth,height=\wxsn\linewidth,keepaspectratio]{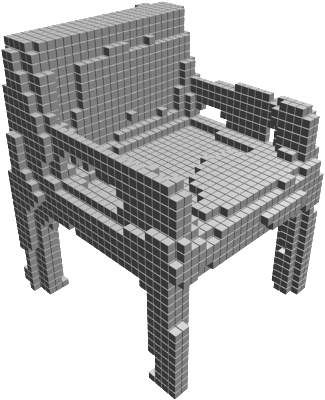} &
  &
  \includegraphics[width=\wxsn\linewidth,height=\wxsn\linewidth,keepaspectratio]{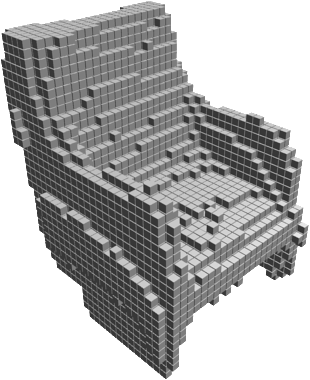} &  
  \includegraphics[width=\wxsn\linewidth,height=\wxsn\linewidth,keepaspectratio]{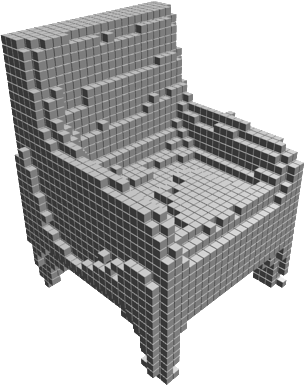} &  
  \includegraphics[width=\wxsn\linewidth,height=\wxsn\linewidth,keepaspectratio]{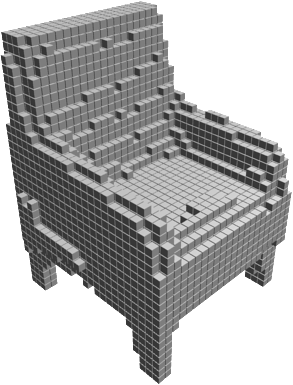} &
  \includegraphics[width=\wxsn\linewidth,height=\wxsn\linewidth,keepaspectratio]{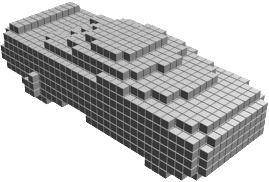} &  
  \includegraphics[width=\wxsn\linewidth,height=\wxsn\linewidth,keepaspectratio]{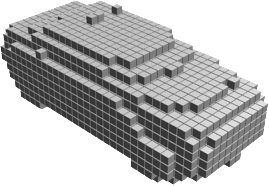} &  
  \includegraphics[width=\wxsn\linewidth,height=\wxsn\linewidth,keepaspectratio]{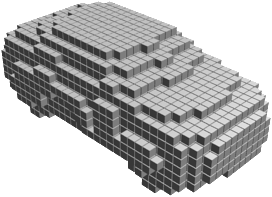} \\

  \hline
  \end{tabular}
  
  \begin{tabular}{ccccccccccccc}
  (b) &\includegraphics[width=\wxsn\linewidth,height=\wxsn\linewidth,keepaspectratio]{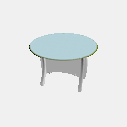} &  
  \includegraphics[width=\wxsn\linewidth,height=\wxsn\linewidth,keepaspectratio]{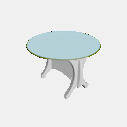} &  
  \includegraphics[width=\wxsn\linewidth,height=\wxsn\linewidth,keepaspectratio]{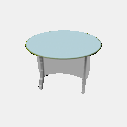} &  
  \includegraphics[width=\wxsn\linewidth,height=\wxsn\linewidth,keepaspectratio]{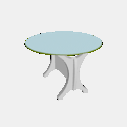} &  
  \includegraphics[width=\wxsn\linewidth,height=\wxsn\linewidth,keepaspectratio]{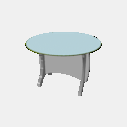} &
  (d) & 
  \includegraphics[width=\wxsnfail\linewidth,height=\wxsnfail\linewidth,keepaspectratio]{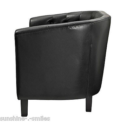} &  
  \includegraphics[width=\wxsnfail\linewidth,height=\wxsnfail\linewidth,keepaspectratio]{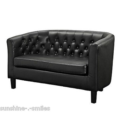} &  
  \includegraphics[width=\wxsnfail\linewidth,height=\wxsnfail\linewidth,keepaspectratio]{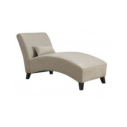} &  
  \includegraphics[width=\wxsnfail\linewidth,height=\wxsnfail\linewidth,keepaspectratio]{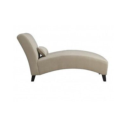} &
  \includegraphics[width=\wxsnfail\linewidth,height=\wxsnfail\linewidth,keepaspectratio]{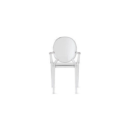} &  
  \includegraphics[width=\wxsnfail\linewidth,height=\wxsnfail\linewidth,keepaspectratio]{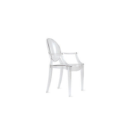} \\
  
  & \includegraphics[width=\wxsn\linewidth,height=\wxsn\linewidth,keepaspectratio]{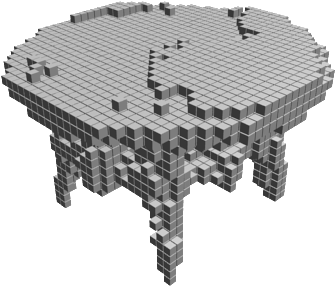} &  
  \includegraphics[width=\wxsn\linewidth,height=\wxsn\linewidth,keepaspectratio]{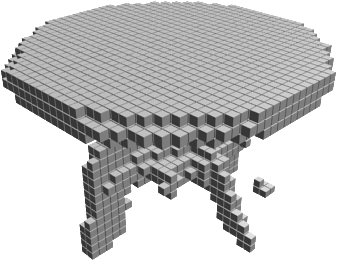} &  
  \includegraphics[width=\wxsn\linewidth,height=\wxsn\linewidth,keepaspectratio]{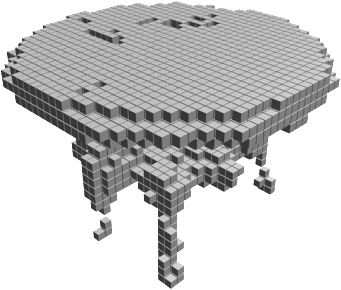} &  
  \includegraphics[width=\wxsn\linewidth,height=\wxsn\linewidth,keepaspectratio]{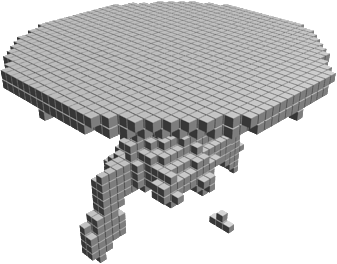} &  
  \includegraphics[width=\wxsn\linewidth,height=\wxsn\linewidth,keepaspectratio]{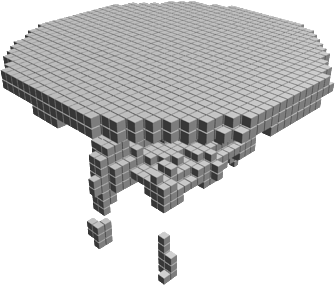} &
  &
  \includegraphics[width=\wxsnfail\linewidth,height=\wxsnfail\linewidth,keepaspectratio]{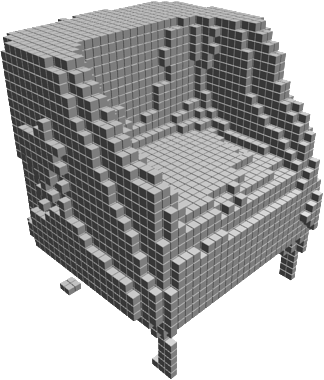} &  
  \includegraphics[width=\wxsnfail\linewidth,height=\wxsnfail\linewidth,keepaspectratio]{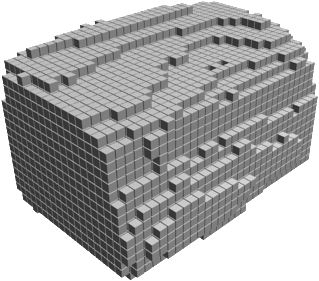} & 
  \includegraphics[width=\wxsnfail\linewidth,height=\wxsnfail\linewidth,keepaspectratio]{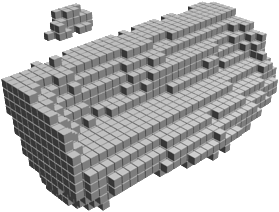} &  
  \includegraphics[width=\wxsnfail\linewidth,height=\wxsnfail\linewidth,keepaspectratio]{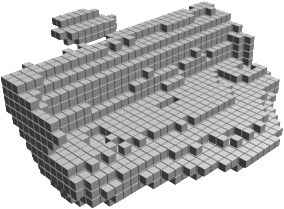} &
  \includegraphics[width=\wxsnfail\linewidth,height=\wxsnfail\linewidth,keepaspectratio]{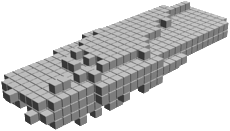} &  
  \includegraphics[width=\wxsnfail\linewidth,height=\wxsnfail\linewidth,keepaspectratio]{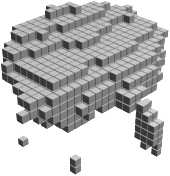}
\end{tabular}
\end{center}
\caption{Sample reconstructions on (a) the ShapeNet~\cite{shapenet} testing set and (c) the Online Products dataset~\cite{ebay}. Top rows are input image sequences (from left to right). Bottom rows are the reconstructions at each time step. (b), (d): Failure cases on each dataset.}
\label{fig:multiview_all}
\vspace{\figurevspace}
\end{figure}

%% file: conclusion.tex
\section{Conclusion}
\label{sec:conclusion}



In this work, we proposed a novel architecture that unifies single- and multi-view 3D reconstruction into a single framework.
Even though our network can take variable length inputs, we demonstrated that it outperforms the method of Kar et al.~\cite{kar2015category} in single-view reconstruction using real-world images. We further tested the network's ability to perform multi-view reconstruction on the ShapeNet dataset~\cite{shapenet} and the Online Products dataset~\cite{ebay}, which showed that the network is able to incrementally improve its reconstructions as it sees more views of an object. 
Lastly, we analyzed the network's performance on multi-view reconstruction, finding that our method can produce accurate reconstructions when techniques such as MVS fail. In summary, our network does not require a minimum number of input images in order to produce a plausible reconstruction and is able to overcome past challenges of dealing with images which have insufficient texture or wide baseline viewpoints.





%% file: acknowledgement.tex
\section{Acknowledgements}

We acknowledge the support of NSF CAREER grant N.1054127 and Toyota Award \#122282. We also thank the Korea Foundation for Advanced Studies and NSF GRFP for their support.

%% file: main.bbl
\begin{thebibliography}{10}

\bibitem{shapenet}
Chang, A.X., Funkhouser, T., Guibas, L., Hanrahan, P., Huang, Q., Li, Z.,
  Savarese, S., Savva, M., Song, S., Su, H., Xiao, J., Yi, L., Yu, F.:
\newblock {ShapeNet: An Information-Rich 3D Model Repository}.
\newblock Technical report, Stanford University --- Princeton University ---
  Toyota Technological Institute at Chicago (2015)

\bibitem{choi2016large}
Choi, S., Zhou, Q.Y., Miller, S., Koltun, V.:
\newblock A large dataset of object scans.
\newblock arXiv preprint arXiv:1602.02481 (2016)

\bibitem{Fitzgibbon1998sfm}
Fitzgibbon, A., Zisserman, A.:
\newblock Automatic 3d model acquisition and generation of new images from
  video sequences.
\newblock In: Signal Processing Conference (EUSIPCO 1998), 9th European, IEEE
  (1998)  1--8

\bibitem{lhuillier2005sfm}
Lhuillier, M., Quan, L.:
\newblock A quasi-dense approach to surface reconstruction from uncalibrated
  images.
\newblock Pattern Analysis and Machine Intelligence, IEEE Transactions on
  \textbf{27}(3) (2005)  418--433

\bibitem{agarwal2009rome}
Agarwal, S., Snavely, N., Simon, I., Seitz, S.M., Szeliski, R.:
\newblock Building rome in a day.
\newblock In: Computer Vision, 2009 IEEE 12th International Conference on, IEEE
  (2009)  72--79

\bibitem{engel2014lsd}
Engel, J., Sch{\"o}ps, T., Cremers, D.:
\newblock Lsd-slam: Large-scale direct monocular slam.
\newblock In: Computer Vision--ECCV 2014.
\newblock Springer (2014)  834--849

\bibitem{haming2010sfmsurvey}
H{\"a}ming, K., Peters, G.:
\newblock The structure-from-motion reconstruction pipeline--a survey with
  focus on short image sequences.
\newblock Kybernetika \textbf{46}(5) (2010)  926--937

\bibitem{fuentes2015slamsurvey}
Fuentes-Pacheco, J., Ruiz-Ascencio, J., Rend{\'o}n-Mancha, J.M.:
\newblock Visual simultaneous localization and mapping: a survey.
\newblock Artificial Intelligence Review \textbf{43}(1) (2015)  55--81

\bibitem{SIFT}
Lowe, D.G.:
\newblock Distinctive image features from scale-invariant keypoints.
\newblock International journal of computer vision \textbf{60}(2) (2004)
  91--110

\bibitem{bhat1998specular}
Bhat, D.N., Nayar, S.K.:
\newblock Ordinal measures for image correspondence.
\newblock Pattern Analysis and Machine Intelligence, IEEE Transactions on
  \textbf{20}(4) (1998)  415--423

\bibitem{saponaro2014sfmtextureless}
Saponaro, P., Sorensen, S., Rhein, S., Mahoney, A.R., Kambhamettu, C.:
\newblock Reconstruction of textureless regions using structure from motion and
  image-based interpolation.
\newblock In: Image Processing (ICIP), 2014 IEEE International Conference on,
  IEEE (2014)  1847--1851

\bibitem{seitz1999voxelcoloring}
Seitz, S.M., Dyer, C.R.:
\newblock Photorealistic scene reconstruction by voxel coloring.
\newblock International Journal of Computer Vision \textbf{35}(2) (1999)
  151--173

\bibitem{kutulako2000spacecarving}
Kutulakos, K.N., Seitz, S.M.:
\newblock A theory of shape by space carving.
\newblock International Journal of Computer Vision \textbf{38}(3) (2000)
  199--218

\bibitem{slabaugh2004voxelcoloring}
Gregory G~Slabaugh, W Bruce~Culbertson, T.M., Stevens, M.R., Schafer, R.W.:
\newblock Methods for volumetric reconstruction of visual scenes.
\newblock International Journal of Computer Vision \textbf{57}(3) (2004)
  179--199

\bibitem{anwar2006voxelcoloring}
Anwar, Z., Ferrie, F.:
\newblock Towards robust voxel-coloring: Handling camera calibration errors and
  partial emptiness of surface voxels.
\newblock In: Proceedings of the 18th International Conference on Pattern
  Recognition - Volume 01. ICPR '06, Washington, DC, USA, IEEE Computer Society
  (2006)  98--102

\bibitem{broadhurst2001probabilistic}
Broadhurst, A., Drummond, T.W., Cipolla, R.:
\newblock A probabilistic framework for space carving.
\newblock In: Computer Vision, 2001. ICCV 2001. Proceedings. Eighth IEEE
  International Conference on. Volume~1., IEEE (2001)  388--393

\bibitem{dame}
Dame, A., Prisacariu, V.A., Ren, C.Y., Reid, I.:
\newblock Dense reconstruction using 3d object shape priors.
\newblock In: Computer Vision and Pattern Recognition (CVPR), 2013 IEEE
  Conference on. (2013)

\bibitem{bao}
Bao, Y., chandraker, M., Lin, Y., Savarese, S.:
\newblock Dense object reconstruction using semantic priors.
\newblock In: Proceedings of the IEEE International Conference on Computer
  Vision and Pattern Recognition. (2013)

\bibitem{lawrence1963single}
Lawrence, G.R.:
\newblock Machine perception of three-dimensional solids.
\newblock Ph. D. Thesis (1963)

\bibitem{nevatia1977single}
Nevatia, R., Binford, T.O.:
\newblock Description and recognition of curved objects.
\newblock Artificial Intelligence \textbf{8}(1) (1977)  77--98

\bibitem{zia2013detailed}
Zia, M.Z., Stark, M., Schiele, B., Schindler, K.:
\newblock Detailed 3d representations for object modeling and recognition,
  TPAMI (2013)

\bibitem{rock2015completing}
Rock, J., Gupta, T., Thorsen, J., Gwak, J., Shin, D., Hoiem, D.:
\newblock Completing 3d object shape from one depth image.
\newblock In: Proceedings of the IEEE Conference on Computer Vision and Pattern
  Recognition. (2015)  2484--2493

\bibitem{choy}
Bongsoo~Choy, C., Stark, M., Corbett-Davies, S., Savarese, S.:
\newblock Enriching object detection with 2d-3d registration and continuous
  viewpoint estimation.
\newblock In: The IEEE Conference on Computer Vision and Pattern Recognition
  (CVPR). (June 2015)

\bibitem{blanz2003face}
Blanz, V., Vetter, T.:
\newblock Face recognition based on fitting a 3d morphable model.
\newblock Pattern Analysis and Machine Intelligence, IEEE Transactions on
  \textbf{25}(9) (2003)  1063--1074

\bibitem{matthews2007face}
Matthews, I., Xiao, J., Baker, S.:
\newblock 2d vs. 3d deformable face models: Representational power,
  construction, and real-time fitting.
\newblock International journal of computer vision \textbf{75}(1) (2007)
  93--113

\bibitem{kemelmacher2011face}
Kemelmacher-Shlizerman, I., Basri, R.:
\newblock 3d face reconstruction from a single image using a single reference
  face shape.
\newblock Pattern Analysis and Machine Intelligence, IEEE Transactions on
  \textbf{33}(2) (2011)  394--405

\bibitem{prisacariu2012accv}
Prisacariu, V.A., Segal, A.V., Reid, I.:
\newblock Simultaneous monocular 2d segmentation, 3d pose recovery and 3d
  reconstruction.
\newblock In: Computer Vision--ACCV 2012.
\newblock Springer (2012)  593--606

\bibitem{sandhu2011nonrigid}
Sandhu, R., Dambreville, S., Yezzi, A., Tannenbaum, A.:
\newblock A nonrigid kernel-based framework for 2d-3d pose estimation and 2d
  image segmentation.
\newblock Pattern Analysis and Machine Intelligence, IEEE Transactions on
  \textbf{33}(6) (2011)  1098--1115

\bibitem{saxena_make3d}
Saxena, A., Sun, M., Ng, A.Y.:
\newblock Make3d: Learning 3d scene structure from a single still image.
\newblock IEEE Trans. Pattern Anal. Mach. Intell. \textbf{31}(5) (may 2009)
  824--840

\bibitem{hoiem2005popup}
Hoiem, D., Efros, A.A., Hebert, M.:
\newblock Automatic photo pop-up.
\newblock ACM transactions on graphics (TOG) \textbf{24}(3) (2005)  577--584

\bibitem{vicente}
Vicente, S., Carreira, J., Agapito, L., Batista, J.:
\newblock Reconstructing pascal voc.
\newblock In: The IEEE Conference on Computer Vision and Pattern Recognition
  (CVPR). (2014)

\bibitem{kar2015category}
Kar, A., Tulsiani, S., Carreira, J., Malik, J.:
\newblock Category-specific object reconstruction from a single image.
\newblock In: Computer Vision and Pattern Recognition (CVPR), 2015 IEEE
  Conference on, IEEE (2015)  1966--1974

\bibitem{lstm}
Hochreiter, S., Schmidhuber, J.:
\newblock Long short-term memory.
\newblock Neural Comput. \textbf{9}(8) (November 1997)  1735--1780

\bibitem{sundermeyer2012lstm}
Sundermeyer, M., Schl{\"u}ter, R., Ney, H.:
\newblock Lstm neural networks for language modeling.
\newblock In: INTERSPEECH. (2012)  194--197

\bibitem{sutskever2014lstm}
Sutskever, I., Vinyals, O., Le, Q.V.:
\newblock Sequence to sequence learning with neural networks.
\newblock In: Advances in neural information processing systems. (2014)
  3104--3112

\bibitem{eigen_depth}
Eigen, D., Puhrsch, C., Fergus, R.:
\newblock Depth map prediction from a single image using a multi-scale deep
  network.
\newblock In: Advances in Neural Information Processing Systems 27.
\newblock (2014)

\bibitem{liu_2015}
Liu, F., Shen, C., Lin, G.:
\newblock Deep convolutional neural fields for depth estimation from a single
  image.
\newblock In: Proc. IEEE Conf. Computer Vision and Pattern Recognition. (2015)

\bibitem{rnn_difficult}
Bengio, Y., Simard, P., Frasconi, P.:
\newblock Learning long-term dependencies with gradient descent is difficult.
\newblock IEEE Transactions on Neural Networks \textbf{5}(2) (Mar 1994)
  157--166

\bibitem{gru}
{Cho}, K., {van Merrienboer}, B., {Gulcehre}, C., {Bahdanau}, D., {Bougares},
  F., {Schwenk}, H., {Bengio}, Y.:
\newblock {Learning Phrase Representations using RNN Encoder-Decoder for
  Statistical Machine Translation}.
\newblock ArXiv e-prints (2014)

\bibitem{resnet}
{He}, K., {Zhang}, X., {Ren}, S., {Sun}, J.:
\newblock {Deep Residual Learning for Image Recognition}.
\newblock ArXiv e-prints (December 2015)

\bibitem{unpooling}
A.Dosovitskiy, J.T.Springenberg, T.Brox:
\newblock Learning to generate chairs with convolutional neural networks.
\newblock In: IEEE International Conference on Computer Vision and Pattern
  Recognition (CVPR). (2015)

\bibitem{everingham2011pascal}
Everingham, M., Van~Gool, L., Williams, C., Winn, J., Zisserman, A.:
\newblock The pascal visual object classes challenge 2012 (2011)

\bibitem{theano}
Bergstra, J., Breuleux, O., Bastien, F., Lamblin, P., Pascanu, R., Desjardins,
  G., Turian, J., Warde-Farley, D., Bengio, Y.:
\newblock Theano: a {CPU} and {GPU} math expression compiler.
\newblock In: Proceedings of the Python for Scientific Computing Conference
  ({SciPy}). (June 2010)

\bibitem{adam}
{Kingma}, D., {Ba}, J.:
\newblock {Adam: A Method for Stochastic Optimization}.
\newblock ArXiv e-prints (2014)

\bibitem{xiang2014beyond}
Xiang, Y., Mottaghi, R., Savarese, S.:
\newblock Beyond pascal: A benchmark for 3d object detection in the wild.
\newblock In: Applications of Computer Vision (WACV), 2014 IEEE Winter
  Conference on, IEEE (2014)  75--82

\bibitem{ebay}
{Song}, H.O., {Xiang}, Y., {Jegelka}, S., {Savarese}, S.:
\newblock Deep metric learning via lifted structured feature embedding.
\newblock ArXiv e-prints (2015)

\bibitem{cgstudio}
:
\newblock Cg studio (2016) [Online; accessed 14-March-2016].

\bibitem{barnes2009patchmatch}
Barnes, C., Shechtman, E., Finkelstein, A., Goldman, D.:
\newblock Patchmatch: A randomized correspondence algorithm for structural
  image editing.
\newblock ACM Transactions on Graphics-TOG \textbf{28}(3) (2009) ~24

\bibitem{mvs}
:
\newblock Openmvs: open multi-view stereo reconstruction library (2015)
  [Online; accessed 14-March-2016].

\bibitem{moulon2013global}
Moulon, P., Monasse, P., Marlet, R.:
\newblock Global fusion of relative motions for robust, accurate and scalable
  structure from motion.
\newblock In: Proceedings of the IEEE International Conference on Computer
  Vision. (2013)  3248--3255

\end{thebibliography}
